\documentclass[sigconf, nonacm]{acmart}

\usepackage{mathrsfs}
\usepackage{booktabs}
\usepackage{array}
\usepackage{multirow}
\usepackage{xspace}
\usepackage[ruled,vlined]{algorithm2e}
\usepackage{bm}

\usepackage{pgfplots}
\usepackage{pgfplotstable}
\usepackage{pgf,tikz}
\pgfplotsset{compat=1.14}

\newcommand{\extdata}[1]{\input{#1}}

\newcommand{\leg}[1]{\addlegendentry{#1}}

\definecolor{greenn}{rgb}{0.30,0.69,0.31}
\definecolor{greennn}{rgb}{0.10,0.50,0.10}
\definecolor{yelloww}{rgb}{1.0000,0.8392,0}
\definecolor{reddd}{rgb}{0.70,0.20,0.20}

\def\cf{\emph{cf.}\xspace}
\def\ie{\emph{i.e.}\xspace}
\def\eg{\emph{e.g.}\xspace}

\def\etal{\emph{et al.}\xspace}
\def\etc{\emph{etc.}\xspace}

\DeclareMathOperator*{\argkmax}{arg\,kmax}

\newcolumntype{L}[1]{>{\raggedright\let\newline\\\arraybackslash\hspace{0pt}}m{#1}}
\newcolumntype{C}[1]{>{\centering\let\newline\\\arraybackslash\hspace{0pt}}m{#1}}
\newcolumntype{R}[1]{>{\raggedleft\let\newline\\\arraybackslash\hspace{0pt}}m{#1}}

\renewcommand{\paragraph}[1]{{\vspace{3pt} \noindent \bf #1}}

\AtBeginDocument{%
  \providecommand\BibTeX{{%
    \normalfont B\kern-0.5em{\scshape i\kern-0.25em b}\kern-0.8em\TeX}}}

\begin{document}

\title{Large-Scale Attribute-Object Compositions}

\author{Filip Radenovic, Animesh Sinha, Albert Gordo, Tamara Berg, Dhruv Mahajan}
\affiliation{%
    \institution{Facebook AI}
    \city{Menlo Park}
    \state{California}
    \country{USA}
}

\renewcommand{\shortauthors}{Radenovic et al.}

\begin{abstract}
We study the problem of learning how to predict attribute-object compositions from images, and its generalization to unseen compositions missing from the training data.
To the best of our knowledge, this is a first large-scale study of this problem, involving hundreds of thousands of compositions.
We train our framework with images from Instagram using hashtags as noisy weak supervision.
We make careful design choices for data collection and modeling, in order to handle noisy annotations and unseen compositions.
Finally, extensive evaluations show that learning to compose classifiers outperforms late fusion of individual attribute and object predictions, especially in the case of unseen attribute-object pairs.
\end{abstract}


\maketitle


%
\begin{figure}
\input{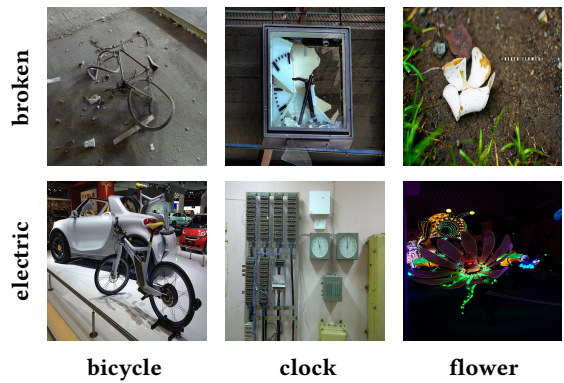}
\vspace{-0.2cm}
\caption{Example attribute-object compositions predicted by our approach in the YFCC100m dataset~\cite{TSF+15}.
\label{fig:teaser}
}
\end{figure}

\section{Introduction}

Attributes are interpretable visual qualities of objects, such as colors, materials, patterns, shapes, sizes, \etc
Their recognition from images has been explored in computer vision, both as specific attributes for faces~\cite{KBN08,LLWT15,WCF16}, people~\cite{DLLT14,WZY+19}, and e-commerce products~\cite{BBS10,HG17}, as well as generic attributes for objects~\cite{LNH09,FEHF09} and scenes~\cite{PH12,LRT+14}.
Attributes have been used to facilitate visual content understanding in applications such as image retrieval~\cite{KBBN11}, product search~\cite{KPG12}, zero-shot object recognition~\cite{ATS16}, image generation~\cite{YYSL16,GCPP20}, \etc

Compared to object classification, where annotating a single dominant object per image is typically enough to train a robust system, attribute prediction requires more complex annotations. An object can often be described by ten or more prominent attributes of different types. For example, a skirt could be \emph{red}, \emph{blue}, \emph{striped}, \emph{long}, and \emph{ruffled}. Other objects may share these same attributes, \eg a pair of pants might be \emph{striped}, \emph{corduroy} and \emph{blue}, suggesting the task of generalizing to new combinations (\emph{blue pants}) from past learnings (\emph{blue skirts}). 
Attribute prediction has been under-explored compared to object classification, and the complexity of obtaining good attribute annotations is likely a factor. Most of the existing labeled attribute datasets lack scale, cover only a narrow set of objects, and are often only partially annotated.

In this work we focus on the problem of joint attribute-object classification, \ie, learning to simultaneously predict not only the objects in an image but also the attributes associated with them, see Figure~\ref{fig:teaser}. 
This poses significant scientific and engineering challenges, mostly due to the large (quadratic) nature of the label space. As an example, 1000 objects and 1000 attributes would already lead to 1M attribute-object combinations. Generating annotations for every combination is therefore impractical, if not infeasible. In addition, some combinations are very rare, and we may not be able to find any training sample for them, even if they may still appear \emph{in the wild}: they are \emph{unseen} at training time but can still arise at inference time in a real-world system.
Most previous works model attributes independently of objects, wishfully hoping that the acquired knowledge about an attribute will transfer to a novel category. 
This reduces the number of training images and annotations required but sacrifices robustness.
On the other hand, work on joint visual modeling of attributes and objects, with analysis on the generalization to unseen combinations, is practically nonexistent.

To address these challenges we propose an end-to-end, weakly-supervised composition framework for attribute-object classification. 
To obtain training data and annotations we build upon the weakly-supervised work of Mahajan~\etal~\cite{MGR+18}, that uses Instagram hashtags as object labels to train the models. 
However, curating a set of hashtag adjectives is a challenging problem by itself, as, unlike nouns, they are mostly non-visual.
We propose an additional hashtag engineering step, in which attributes are selected in a semi-automatic manner to fit the desired properties of visualness, sharedness across objects, and interpretability.
Our final training dataset consists of $78M$ images spanning $7694$ objects, $1237$ attributes, and $280k$ compositions with at least $100$ occurrences, which is significantly larger than any attribute-based dataset in number of images and categories.

We also propose a multi-head architecture with three output classifier heads: (i) object classifiers; (ii) object-agnostic attribute classifiers; and, (iii) attribute-object classifiers.
Instead of explicitly learning a linear classifier for every attribute-object fine-grain combination -- which has high requirements in terms of memory and computation, and has limited generalization-- and motivated by the work of Misra~\etal~\cite{MGH17}, we incorporate a module that composes these classifiers by directly reasoning about them in the classifier space.
This composition module takes the object and attribute classifier weights from the network and learns how to compose them into  attribute-object classifiers. Crucially, this allows the model to predict, at inference time, combinations not seen during training. 

We show that a vanilla implementation of the composition network has performance and scalability issues when trained on our noisy dataset with missing object and attribute labels. Hence, we propose a set of crucial design changes to make the composition network effective in our real-world setting. These choices include a more suitable loss function that caters better to dataset noise and a selection strategy to reduce the set of candidate attribute-object combinations considered during training and inference.

Finally, we extensively evaluate our method at different data scales by testing the performance on both seen and unseen compositions and show the benefits of explicitly learning compositionality instead of using a late fusion of individual object and attribute predictions.
Additionally, to advocate the usage of the proposed framework beyond weakly-supervised learning, we evaluate its effects on our internal Marketplace dataset with cleaner labels.


\section{Related Work}
\label{sec:related}

\paragraph{Object classification and weak supervision.} 
Object classification and convolutional neural networks attracted a lot of attention after the success of Krizhevsky~\etal~\cite{KSH12}.
Their success is mainly due to the use of very large annotated datasets, \eg ImageNet~\cite{RDS+15}, where the acquisition of the training data requires a costly process of manual annotation.
Training convolutional networks on a very large set of weakly-supervised images by defining the proxy tasks using the associated meta-data has shown additional benefits for the image-classification task~\cite{DWP+15,JVJV16,GRS17,SSSG17,LJJV17,MGR+18,VNBV18}.
Some examples of proxy tasks are hashtag prediction~\cite{DWP+15,GRS17,MGR+18,VNBV18}, search query prediction~\cite{SSSG17}, and word n-gram prediction~\cite{JVJV16,LJJV17}.

Our approach builds upon the work of Mahajan~\etal~\cite{MGR+18} which learns to predict hashtags on social media images, after a filtering procedure that matches hashtags to noun synsets in WordNet~\cite{M95}.
We extend these noun synsets 
 with adjective synsets corresponding to visual attributes.
Unlike object-related nouns, that are mostly visual, most of the attributes selected in this a manner were non-visual, and required us to apply an additional cleaning procedure.

\paragraph{Visual attribute classification.}
We follow the definition of visual attributes by Duan~\etal~\cite{DPCG12}: 
\emph{Attributes are visual concepts that can be detected by machines, understood by humans, and shared across categories.}
The mainstream approach to learn attributes is very similar to the approach used to learn object classes: training a convolutional neural network with discriminative classifiers and carefully annotated image datasets~\cite{LLWT15,WCF16,SL16,SZX+16,LKZ+17}.
Furthermore, labeled attribute image datasets either lack the data scale~\cite{KBBN09,LNH09,RF10,PH12,CGG12,ILA15,PH16,ZFL+19} common to the object datasets, contain a small number of generic attributes~\cite{LNH09,RF10}, and/or cover few specific categories such as \emph{person}~\cite{LKS11,SJ11}, \emph{faces}~\cite{KBBN09,LLWT15}, \emph{clothes}~\cite{CGG12,YG14,LLQ+16}, \emph{animals}~\cite{LNH09}, \emph{scenes}~\cite{PH12}.
In this work, we explore learning diverse attributes and objects from large-scale weakly-supervised datasets.

\paragraph{Composition classification.}
The basic idea of compositionality is that new concepts can be constructed by combining the primitive ones.
This idea was previously explored in natural language processing~\cite{ML08,BZ10,G10,SPW+13,NLB14}, and more recently in vision~\cite{SF11,PSB16,ZKCC17,CG14,MGH17,NG18,SFCG18}.
Compositionality in vision can be grouped into following modeling paradigms: object-object (noun-noun) combinations~\cite{PSB16}, object-action-object (noun-verb-noun) interactions~\cite{SF11,ZKCC17},  attribute-object (adjective-noun) combinations~\cite{CG14,MGH17,NG18}, and complex logical expressions of attributes and objects~\cite{SFCG18,CNJZ20}. In this work, we focus on the attribute-object compositionality.

Prior work also focuses on the \emph{unseen} compositions paradigm~\cite{CG14,MGH17,NG18}, where a part of the composition space is \emph{seen} at the training time, while new \emph{unseen} compositions appear at inference, as well.
Towards that end, Chen~and~Grauman~\cite{CG14} employ tensor completion to recover the latent factors for the 3D attribute-object tensor, and use them to represent the unobserved classifier parameters.
Misra~\etal~\cite{MGH17} combine pre-trained linear classifier weights (vectors) into a new compositional classifier, using a multilinear perceptron (MLP) which is trained with seen compositions but shows generalization abilities to unseen ones.
Finally, Nagarajan~and~Grauman~\cite{NG18} model attribute-object composition as an attribute-specific invertible transformation (matrix) on object vectors.

Motivated by the idea from Misra~\etal~\cite{MGH17}, we combine attribute and object classifier weights with an MLP to produce composition classifiers. 
Unlike~\cite{MGH17}, we learn these constituent classifiers in a joint end-to-end pipeline together with image features and compositional MLP network. As discussed during the introduction, further design changes (\eg changes in the loss and the composition selection) are also required in a large-scale, weakly-supervised setting. 


\section{Modeling}
\label{sec:modeling}

In this section we first describe the data collection process to create our datasets.
  Then, we discuss the full pipeline architecture and loss functions employed to jointly train it in an end-to-end fashion.
Finally, we describe our efficient inference procedure, which does not require  computing predictions for all the compositions.

\subsection{Training and Evaluation Data}

\subsubsection{Instagram Datasets}
We follow the data collection pipeline of~\cite{MGR+18}, extended for the purpose of collecting attribute hashtags in addition to object hashtags.
This simple procedure collects public images from Instagram\footnote{\url{https://www.instagram.com}} after matching their corresponding hashtags to WordNet synsets~\cite{M95}:
(i) We select a set of hashtags corresponding to noun synsets for the objects, and adjective synsets for the attributes.
(ii) To better fit the compositional classification task, we download images that are tagged with at least one hashtag from the object set and at least one hashtag from the attribute set.
(iii) Next, we apply a hashtag deduplication procedure~\cite{MGR+18}, that utilizes WordNet synsets~\cite{M95} to merge multiple hashtags with the same meaning into a single canonical form (\eg, \texttt{\#brownbear} and \texttt{\#ursusarctos} are merged). 
In addition, for adjectives only, we merge relative attributes into a single canonical form, as well, (\eg, \texttt{\#small}, \texttt{\#smaller}, \texttt{\#smallest}).
(iv) Finally, for each downloaded image, each hashtag is replaced with its canonical form.
The canonical hashtags for objects, attributes, and their pairwise compositions are used as label sets for training and inference.

\paragraph{Attribute visualness, sharedness, and interpretability.}
Unlike object (noun) classes, that are mostly visual, attribute (adjective) classes are often non-visual and tend to be noisier.
In addition to the hashtag filtering applied in~\cite{MGR+18}, we apply two automatic strategies to clean them.
These strategies are inspired by the attribute definition, \ie, we would like them to be \emph{recognizable by computer vision} (visualness), and they should be \emph{shared across objects} (sharednesss).

We implement a similar strategy as in~\cite{BBS10} to generate a visualness score for each attribute: 
(i) We start by training linear classifiers for all attributes on top of image features from~\cite{MGR+18}. 
(ii) We then evaluate precision@5 for each attribute on a held out validation set, and use it as a visualness score.
Examples of attributes that have low visualness score: \texttt{\#inspired}, \texttt{\#talented}, \texttt{\#firsthand}, \texttt{\#atheist}.

To evaluate attribute sharedness across objects, we analyse their co-occurrence statistics. 
For a given attribute, sharedness score is defined as the number of objects it occurs more than 100 times with, weighted by the logarithm of inverse attribute frequency.
These scores are finally normalized to $[0, 1]$ range, to be comparable with visualness score.
For example, \texttt{\#aerodynamic} attribute has a high visualness score, but occurs exclusively with \texttt{\#airplane} category, thus having a low sharedness score.

Finally, we rank attributes based on the product of their visualness and sharedness score, and manually select those that are \emph{interpretable by humans} based on a small set of random images associated with their respective hashtags.
This is done to make sure visual attributes are representing object features that a user would use to express themselves in, for example, product search.
The filtering procedure is lightweight, as the head of the ranked list already provides a very clean set of attributes. 
For reference, the full list of adjective hashtags contains 10k entries, and only 1237 are selected to satisfy all three properties.
By combining attributes with the objects they describe, we create the following two datasets of different scale, to be used in the experimental analysis.

\paragraph{IG-504-144.}
A dataset with 504 object and 144 attribute categories.
It contains 8904 attribute-object compositions with at least 100 occurrences. 
In order to evaluate the \emph{unseen} scenario, we randomly split the compositions with a 20/80\% ratio, \ie, 1729 (20\%) compositions are selected as \emph{unseen} and 7175 (80\%) as \emph{seen}.
We then label all images that contain at least one of unseen compositions as \emph{unseen}, and the rest as \emph{seen}.
The train partition contains 2.5M images selected from the seen image split only.
The test partition contains 740k images selected from both seen and unseen splits.

\paragraph{IG-8k-1k.}
A dataset with 7694 object and 1237 attribute categories.
It contains more than 280k attribute-object compositions with at least 100 occurrences. 
In this case, we split the compositions with a 30/70\% ratio, \ie, 83461 (30\%) compositions are selected as \emph{unseen} and 196646 (70\%) as \emph{seen}.
The train partition contains 78M images images selected from the seen image split only.
The test partition contains 890k images selected from both seen and unseen splits.

This dataset is significantly larger than any other publicly available dataset containing attributes, both in the number of images and class set size, while covering a wide variety of object categories. Note that our train and test datasets are weakly-supervised, and thus suffer from a considerable amount of noise.

\subsubsection{Marketplace Dataset}
To verify that our weakly-supervised evaluation translates to the fully-supervised scenario, we also evaluate our method on the internal Marketplace\footnote{\url{https://www.facebook.com/marketplace}} dataset. This dataset is smaller in scale than our largest Instagram dataset, but the test data is collected in a fully-supervised setting with human supervision.

We leverage the Marketplace C2C (customer-to-customer) image dataset, which contains public images uploaded by users on the platform for selling their product items to other users. We follow a two-level taxonomy where the first level describes the attribute types (\emph{color}, \emph{pattern}, \emph{embellishments}, \emph{etc}) and the leaf level describes the attribute values (\emph{red}, \emph{blue}, \emph{polka dot}, \emph{fringe}, \emph{etc}). This dataset contains 6M train and 238k test images spanning across 992 product categories, 22 attribute types, 672 attribute values, and 39k compositions. We collected annotations across 4 commerce verticals (Clothing, Accessories, Motors and Home \& Garden). 

We manually annotated around 2M images to construct our seed training dataset. We then augmented our training dataset by running an existing model to mine more positive annotations from millions of unlabeled C2C images and added around 4M more images to our existing training set, making it 6M in total. All images in the test set were manually annotated by human raters and do not contain any model generated annotations. We carefully sampled images across all geographies to reduce cultural bias in categories like clothing. We also tried to reduce the attribute-attribute co-occurrence bias during the image sampling. For instance, we observed that \emph{camouflage pants} mostly occurred with \emph{green pants}. Finally, we randomly selected $50\%$ of the compositions as unseen and remove the corresponding images from the training set.

\begin{figure*}[t]
    \centering
    \includegraphics[width=0.85\textwidth]{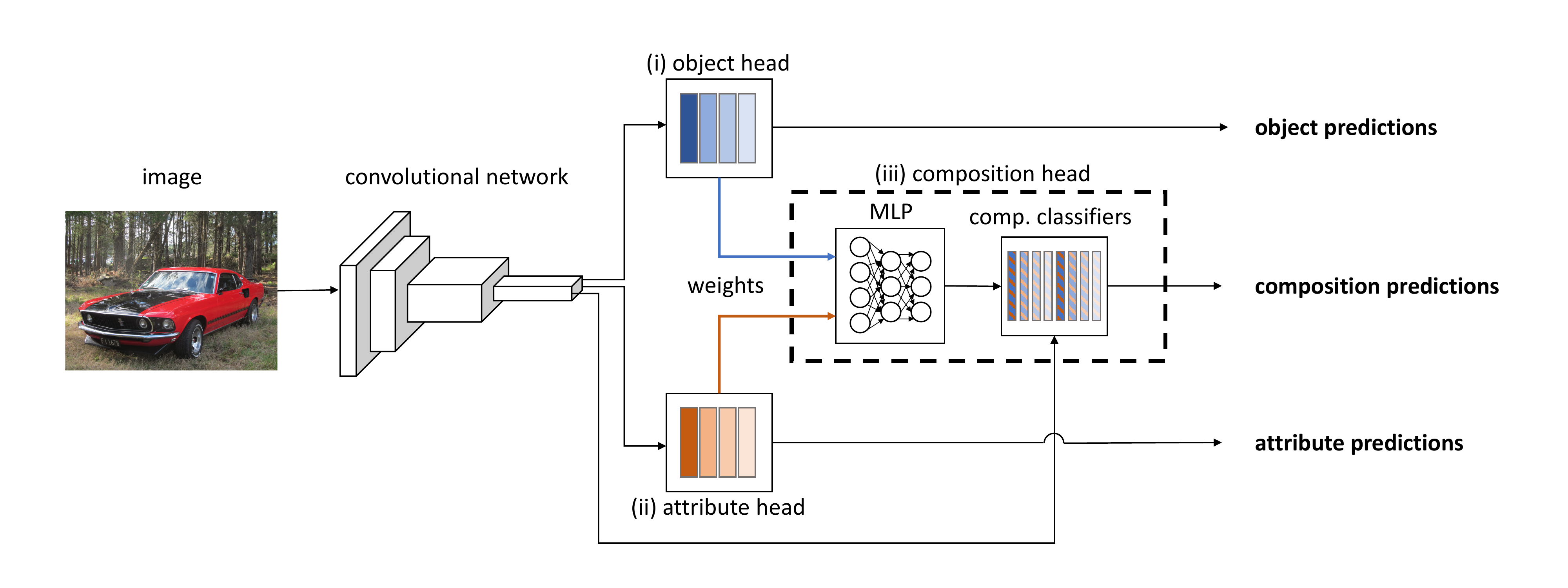}
    \vspace{-0.5cm}
    \caption{CompNet: Our proposed architecture with three output streams: (i) object predictions; (ii) object-agnostic attribute predictions; and, (iii) attribute-object composition predictions. Composition head takes individual object and attribute classifier weights as inputs and produces a composition classifier for each pair, which is then applied to image features.}
    \label{fig:pipeline}
\end{figure*}

\subsection{Pipeline Architecture}
Our CompNet pipeline consists of a convolutional neural network feature generator, followed by three heads, one for each task: object, attribute, and composition classification.
An overview of the pipeline is depicted in Figure~\ref{fig:pipeline}. 
Object and attribute heads are both single fully-connected layers, trained on their respective hashtag sets, \ie, objects and attributes. The score of the attributes ($s_a$) and objects ($s_b$) is computed as the dot product between the image features and the linear classifiers.

\paragraph{Composition head.}
We adopt the approach of Misra~\etal~\cite{MGH17} to compose complex visual classifiers from two classifiers of different types (attribute and object).
For example, given the classifier weights of attribute \texttt{\#red} and object \texttt{\#car}, this method outputs classifier weights for their composition \texttt{\#red\_car} (\cf Figure~\ref{fig:pipeline}.)

Let $\bm{w}_a$ and $\bm{w}_o$ denote the $D$-dimensional linear classifier vectors for attribute $a$ and object $o$, that are applied to the $D$-dimensional image features $\phi(I)$ extracted from a Convolutional Neural Network (CNN) 
\footnote{In our implementation we also have a bias term in the classifier, which we omit from the text for better readability. To integrate it we augment the $\bm{w}$ classifiers with the bias term and the feature $\phi(I)$ with a bias multiplier (set to 1) as an additional dimension.}.
Next, let us denote set of all attributes as $\mathcal{A}$, and all objects as $\mathcal{O}$.
Attribute-object pairs are composed into the composition $ao$, or more precisely, their classifier vectors $\bm{w}_a$ and $\bm{w}_o$ are composed into the composition classifier vector $\bm{w}_{ao}$ of the same size.
The composition is performed by feeding a concatenated pair of vectors $(\bm{w}_a, \bm{w}_o)$ into the multi-layer perceptron (MLP) that outputs a vector $\bm{w}_{ao}$.
Formally, for each $a \in \mathcal{A}$ and each $o \in \mathcal{O}$, the classifier for composition $ao$ is computed as
\begin{equation}
    \bm{w}_{ao} = \mathscr{C}(\bm{w}_a, \bm{w}_o),
\end{equation}
where $\mathscr{C}$ is a composition function parameterized with an MLP and learned from our training data.
Finally, $\bm{w}_{ao}$ is applied on the image feature $\phi(I)$,
\begin{equation}
    \label{eq:logit}
    s_{ao} = \bm{w}_{ao} \cdot \phi(I),
\end{equation}
where $\cdot$ denotes the dot-product and $s_{ao}$ is the attribute-object logit.                                                             

We now discuss how we diverge from Misra~\etal~\cite{MGH17}, by proposing a novel composition loss function better suited for the weakly-supervised scenario (see Section~\ref{sec:loss}), and the strategy for efficient inference in large attribute-object spaces (see Section~\ref{sec:inference}). 

\subsection{Loss Functions}
\label{sec:loss}

Our pipeline unifies three tasks into a single architecture: object, attribute, and composition classification.
The final loss is a weighted sum of loss functions for each task:
\begin{align}
\label{eq:final_loss}
L(I, y_a, y_o;\phi , \bm{w}_a, \bm{w}_o, \mathscr{C}) &= 
\lambda_a L_a\left(s_a, \; y_a\right) 
+ \lambda_o L_o\left(s_o, \; y_o\right) + \nonumber \\
& \hspace{1cm} + \lambda_{ao} L_{ao}\left(s_{ao}, \; y_{a} \wedge y_{o}\right),
\end{align}
where $y_{a}$ and $y_{o}$ are attribute and object labels, respectively.

\subsubsection{Object and attribute loss}
\label{sec:obj_attr_losss}
For the object and attribute loss functions ($L_o$ and $L_a$, respectively) we use the standard cross-entropy loss~\cite{GBC16} adjusted for the multi-label scenario~\cite{MGR+18}, after applying softmax to the output.
The multi-label adjustment is needed because images often contain multiple object and attribute hashtags.
Each positive target is set to be $1/k$, where $k \geq 1$ corresponds to the number of hashtags from that specific task (object or attribute).

\subsubsection{Composition loss}
\label{sec:comploss}
In~\cite{MGH17}, the authors propose to use the binary cross-entropy loss, after applying a sigmoid on the logit output from (\ref{eq:logit}), in order to get a probability score:
\begin{align}
    p_{ao} &= \text{sigmoid} (s_{ao}), \nonumber \\
    L_{ao} &= y_{ao} \log (p_{ao}) + (1 - y_{ao}) \log (1 - p_{ao}),
\end{align}
where the label $y_{ao}$ is $1$ only if the image has the composition $ao$ present, \ie $y_{a}=1 \wedge y_{o}=1$.
Unfortunately, in our experiments, we obtained results that are significantly worse than a simple Softmax Product baseline (discussed in Section~\ref{sec:experiments}).
We believe this is due to training with incomplete noisy annotation, consisting of weak positive labels and no negative ones.
These findings coincide with observations in other weakly-supervised approaches exploring binary cross-entropy~\cite{JVJV16,SSSG17,MGR+18}.

However, naively applying a softmax on all compositions, as we did for the individual objects and attributes, is prohibitively expensive, as we have millions of compositions in our larger training dataset, or hundreds of thousands if we consider only those that have at least 100 occurrences.
The computation cost is not only in back-propagating the loss across all compositions, but also due to the computation of composition classifiers.
Instead, we rely on an efficient softmax approximation. Contrary to previous approximations, that \eg simply update the classes present in the training batch~\cite{JVJV16}, we leverage the scores of the individual object and attribute scores to construct the set of hard negative classes.

\paragraph{Hard negative composition classes.}
Let us define probability of attribute $a$ as $p_a$, and of object $o$ as $p_o$, which we obtain as the softmax output of individual object and attribute heads of the pipeline.
Additionally, let us define a set of compositions present in the training set (\textit{seen} only) as $\mathcal{C}^{\text{seen}}$.
For each image, we find a set of $k$ negatives $\mathcal{N}$ to be used in softmax computation,
\begin{align}
    \mathcal{N} = \argkmax_{a'o' \in \mathcal{C}^{\text{seen}} \setminus \{ao \in I\}} (p_{a'} p_{o'}),
    \label{eq:hard_neg}
\end{align}
where $\{ao \in I\}$ represent all the positive compositions for the image, computed as the Cartesian product of the positive object and positive attributes.

In other words, \emph{hard} negatives are chosen for each image based on the individual softmax probability product, which is an approximation of the joint attribute and object probability.
Next, we compute the composition classifiers for each positive $ao$ and the set of negatives $\mathcal{N}$, and get the logit scores after applying them to the image features, respectively $s_{ao}$ and $\mathcal{S}_{ao}^{\mathcal{N}}$.

The approximate softmax joint attribute-object probability for a composition $ao$ is now:
\begin{align}
    p_{ao} = \text{softmax}_{ao} \left( \{s_{ao}\} \cup \mathcal{S}_{ao}^{\mathcal{N}} \right),
    \label{eq:joint_prob}
\end{align}
where $\text{softmax}_{ao}$ denotes the value associated with $ao$ in the softmax vector.  
Note that, in a naive approximation of the softmax, one would use entire space of compositions $\mathcal{C}$ to select hard negatives, instead of only using seen ones $\mathcal{C}^{\text{seen}}$ as in~(\ref{eq:hard_neg}).
Given that the majority of the compositions are actually not present in the training data, the unseen ones are often selected as negatives (never as positives).
We observe drastically worsened performance on unseen compositions in that case, see ablation in Section~\ref{sec:model_study} for details.

\paragraph{Conditional probability term.}
When searching for hard negative compositions in the weakly-supervised setup, false negatives are often selected.
To alleviate this, we additionally use the conditional rule to approximate the joint probability. 
As an example, if the positive composition is \texttt{\#red\_dress}, there is a high probability that other \texttt{\#red} objects are negative examples for that image (\eg, \texttt{\#red\_car} and \texttt{\#red\_chair}), so we compute the joint attribute-object probability through the conditional probability of the object given attribute.
Formally:
\begin{align}
    \label{eq:cond_prob}
    p_{ao}^a = p_a \text{softmax}_{ao} \left( \left\{ s_{a'o'} \mid a'=a \land o' \in \mathcal{O} \right\} \right).
\end{align}
Similarly, we can compute $p_{ao}^o$ through conditional probability of the attribute given object.

\begin{figure}[t]
    \centering
    \includegraphics[width=0.4\textwidth]{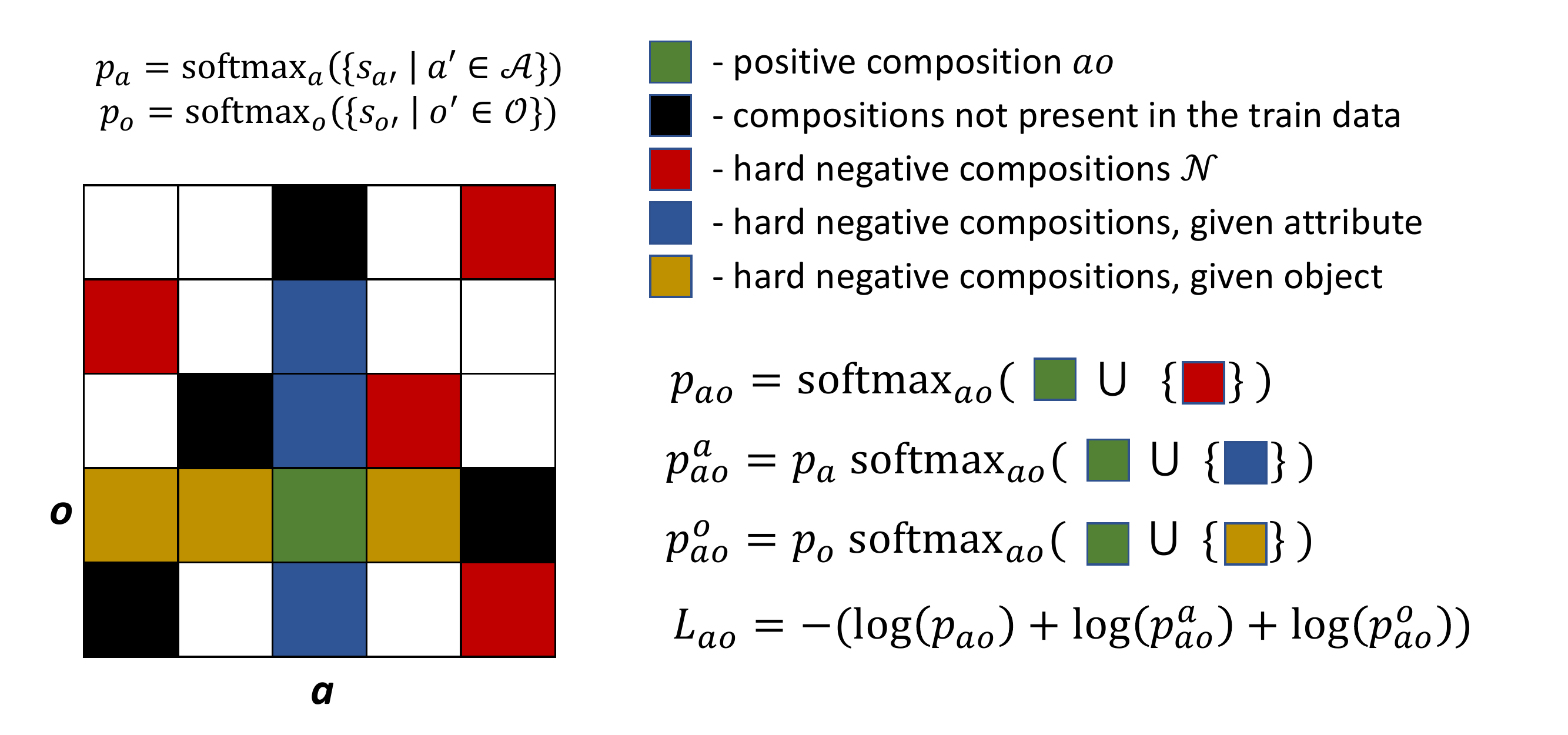}
    \vspace{-0.5cm}
    \caption{Illustration of our proposed composition loss. Joint attribute-object probability is approximated using hard negatives of the entire composition space, and through conditional probabilities, \ie, by fixing attribute and object class.}
    \label{fig:loss}
\end{figure}

The final composition loss now becomes
\begin{align}
    L_{ao} = - \left( \log(p_{ao}) + \log(p_{ao}^a) + \log(p_{ao}^o) \right).
    \label{eq:comploss}
\end{align}
A simplified illustration of this procedure is depicted in Figure~\ref{fig:loss}.

It is worth mentioning that there is often more than one positive composition per image. In fact, each pairwise combination of positive object and attribute hashtags is assumed to be a positive composition.
We additionally use a multi-label version of the loss by weighting each positive label contribution to the loss with $1/k$, where $k\geq1$ is the number of all pairwise combinations of attribute-object hashtags present in the image.
A step-by-step procedure for the loss computation is summarized in Algorithm~\ref{alg:loss}.

\begin{algorithm}[t]
\caption{CompNet loss computation.}
\label{alg:loss}
 \DontPrintSemicolon
 \SetAlgoLined
 \textbf{input:}\\ 
 \hspace{0.38cm} -- batch of images $B$\\
 \hspace{0.38cm} -- feature generator $\phi$\\
 \hspace{0.38cm} -- object classifiers $\left\{\bm{w}_o' \mid o' \in \mathcal{O} \right\}$\\
 \hspace{0.38cm} -- attribute classifiers $\left\{\bm{w}_a' \mid a' \in \mathcal{A} \right\}$\\
 \hspace{0.38cm} -- composition MLP network $\mathscr{C}$\\
 \hspace{0.38cm} -- object and attribute \texttt{\#hashtags}\\
 \For{\upshape each image $I$ in $B$}{
    -- compute loss for object and attribute head (Sec.~\ref{sec:obj_attr_losss})\\
    // we use $p_a$ and $p_o$ from prev. step for the following \\
    -- select hard negative compositions (Eq.~\ref{eq:hard_neg})\\
    -- compute approx. composition probability $p_{ao}$ (Eq.~\ref{eq:joint_prob})\\
    -- compute approx. comp. probability $p_{ao}^a$ and $p_{ao}^o$ (Eq.~\ref{eq:cond_prob})\\
    -- compute composition loss using $p_{ao}$, $p_{ao}^a$, $p_{ao}^o$ (Eq.~\ref{eq:comploss})\\
    -- final loss is a weighted sum of three losses (Eq.~\ref{eq:final_loss})\\
 }
\end{algorithm}

\subsection{Inference}
\label{sec:inference}
Outputting all possible composition scores for each image is computationally expensive, even at inference. 
In~\cite{MGH17} scores are computed only for a predefined set of \emph{unseen} compositions.
Practically, this is unrealistic, as we do not have an \emph{oracle} to decide which compositions will be useful at inference.
More realistically, in~\cite{NG18} scores are computed for a larger predefined set of \emph{seen} and \emph{unseen} compositions.
However, that still leaves out the majority of compositions, which simplifies inference, and diverges from realistically deployed system.
In our setup, we do not assume any predefined set of compositions and propose a simple inference strategy: compute composition scores on a shortlist, consisting of every pairwise combination of top-$k_a$ attributes and top-$k_o$ objects predicted by attribute and object classifiers individually.
Thus, the final composition logit output will have $k_a \times k_o$ entries, and the probabilities are computed by applying a softmax on these logits.
If the composition being evaluated is not present in this shortlist, its probability is considered as 0 for the performance computation.


\begin{table*}[t]
\caption{
Training implementation details. 
GPUs: total number of GPUs across machines with 8 GPUs per machine;
Batch: total batch size, each GPU processes 32 images at a time and batch normalization (BN)~\cite{IS15} statistics are computed on these 32 image sets;
Warm-up: to set the learning rate, we follow the linear scaling rule with gradual warm-up~\cite{GDG+17} during the first X\% of training iterations; 
LR init: learning rate initialization, this number is additionally multiplied with the total batch size. 
}
\vspace{-0.2cm}
\centering
\small
\def\arraystretch{0.8} 
\def\cw{1cm}
\begin{tabular}{L{1.5cm}C{0.8cm}C{0.8cm}C{0.9cm}C{1.5cm}C{1.5cm}C{1.8cm}C{1.7cm}C{1cm}C{1.5cm}}
\toprule
Dataset & GPUs & Batch & Epochs & Warm-up & LR init & LR schedule & $\left[\lambda_a,\lambda_o,\lambda_{ao}\right]$ & $|\mathcal{N}|$ & top-$k_a{\times}k_o$ \\
\midrule
IG-8k-1k & 128 & 4096 & 40 & 5\% linear & 0.1/256 & ${\times}$0.5 in 10 steps & $\left[1, 1, 1 \right]$ & 10000 & 100${\times}$100 \\
IG-504-144 & 128 & 4096 & 60 & 5\% linear & 0.1/256 & ${\times}$0.5 in 10 steps & $\left[1, 1, 1 \right]$ & 5000 & 50${\times}$50 \\
Marketplace & 128 & 4096 & 32 & 12.5\% linear & 0.001/256 & cosine~\cite{LH16} & $\left[1, 1, 10 \right]$ & all seen & 100${\times}$100 \\
\bottomrule
\end{tabular}
\label{tab:params}
\end{table*}

\section{Experiments}
\label{sec:experiments}
In this section we discuss implementation details, study different components of our method, and compare to the baselines.

\subsection{Training and evaluation setup}
\paragraph{Composition head details.}
The function $\mathscr{C}$ is parametrized as a feed-forward multi-layer perceptron (MLP) with 2 hidden layers and a dropout~\cite{SHK+14} rate of $0.3$. Following~\cite{MGH17}, we use leaky ReLU~\cite{MHN13} with coefficient $a=0.1$. 
The input to the MLP is a $2D$ dimensional vector constructed by concatenation of the attribute and object classifiers. Both hidden layers are $D$ dimensional and the final output is a $D$ dimensional attribute-object classifier.

\paragraph{Instagram details.}
We use a ResNeXt-101~$32{\times}4$~\cite{XGD+17} network trained from scratch by synchronous stochastic gradient descent (SGD).
The final hyper-parameters are detailed in Table~\ref{tab:params}, and used throughout the experiments unless explicitly stated.
Training on IG-8k-1k 78M images for 60 epochs took ${\sim}15$ days.
Object and attribute tasks are evaluated using precision@1 (P@1), \ie, the percentage of images for which the top-scoring prediction is correct. Attribute-object pairs
are evaluated using mean average precision (mAP), \ie, for each composition, rank all images and compute average precision, averaged across \emph{seen} (S) and \emph{unseen} (U) composition splits separately. 
We would like to point out that unlike many previous approaches, we do not have separate test data for seen and unseen compositions. This makes the unseen evaluation more challenging since training is biased towards seen attribute-object pairs. However, this mimicks the realistic deployment scenario where we can not distinguish between the seen and unseen attribute-object pairs.

\begin{figure}[t]
    \centering
    \includegraphics[width=0.35\textwidth]{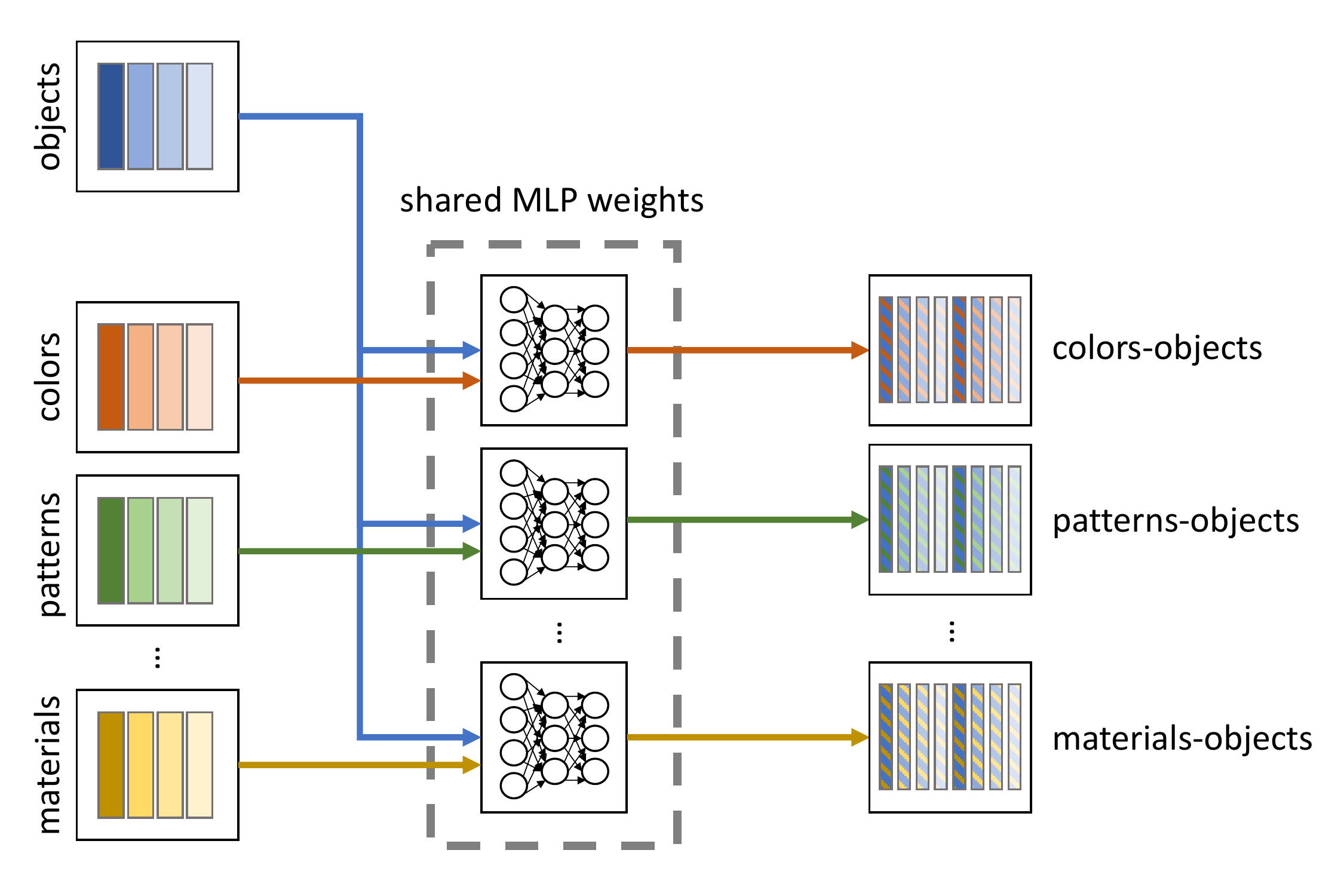}
    \caption{Composing classifiers for Marketplace. Each attribute type is composed with objects separately to generate 22 attribute-object heads with different attribute types.}
    \label{fig:marketplace}
\end{figure}

\paragraph{Marketplace details.}
We build on top of a previous system that was trained on our internal Marketplace attributes dataset. 
The trunk of the pipeline is a ResNet-50~\cite{HZRS16} pre-trained on hashtags following~\cite{MGR+18}.
Attributes are split into 22 types, and a separate head is trained for each type.
We extend the pipeline by adding an object category head with the 992 products, and combine it with each of the 22 attribute heads to create 22 composition heads, all of them sharing their weights, see Figure~\ref{fig:marketplace}.
The final hyper-parameters are detailed in Table~\ref{tab:params}. 
Due to the multi-head nature of attributes for this dataset, hard-negatives are selected for each attribute type separately.
Object performance is evaluated using P@1, while attribute and composition performance using mAP.

\subsection{Model study}
\label{sec:model_study}

\begin{table}[t]
\caption{Model study of our proposed CompNet method. Performance is reported as mAP averaged over seen (S) and unseen (U) composition classes from IG-504-144.
}
\centering
\def\arraystretch{0.8} 
\def\cw{0.55cm}
\begin{tabular}{L{0.45cm}L{5.5cm}C{\cw}C{\cw}}
\toprule
& \multirow{2}{*}{CompNet} & \multicolumn{2}{c}{IG-504-144} \\
\cmidrule(lr){3-4}
&& S & U \\
\midrule
(i) & vanilla softmax approximation & 18.6 & 5.3 \\
\midrule
(ii) & remove unseen from hard negatives & 19.1 & 10.3 \\
\midrule
(iii) & add conditional probability terms & 19.3 & 11.2 \\
\bottomrule
\end{tabular}
\label{tab:ablation}
\end{table}

To understand the effect of different design choices we experiment on the IG-504-144 dataset. Table~\ref{tab:ablation} shows the performance of our proposed composition approach, quantifying the benefits of our proposed modifications to the loss function. Row (i) shows the results with vanilla softmax approximation where we do not remove unseen compositions from the hard negatives and remove conditional probability terms (\cf  Section~\ref{sec:comploss}). The model learns to output low scores for unseen compositions, as it only uses them as negatives at training. In row (ii) we remove such compositions from the hard negatives (Eq.~(\ref{eq:hard_neg})) and observe a significant improvement in the performance of unseen attribute-object pairs ($5.3$ vs. $10.3$). Finally, in row (iii) we add the conditional terms back (Eq.~\ref{eq:comploss}) and observe a further improvement of $0.9$ mAP. Note that for the seen attribute-object pairs our design choices achieve a moderate improvement of $0.7$ mAP ($18.6$ vs.$19.3$) compared to the $5.9$ mAP improvement ($5.3$ vs. $11.2$) for the unseen ones.

We evaluate the robustness of our pipeline to the choice of top-$k$ hard negatives (Eq.~\ref{eq:hard_neg}) during training, and to the choice of output size top-$k_a{\times}k_o$ (Section~\ref{sec:inference}) during inference.
Results are presented in Table~\ref{tab:output_size} for unseen compositions. First let us consider each row. We observe that performance is robust to the number of hard negatives during training. There is less than $0.15$ mAP degradation when we increase the hard negatives from $900$ to using all seen compositions. Now let us consider each column where we increase the number of compositions evaluated during inference. We observe an improvement of around $0.9$ mAP from  top-$30{\times}30$ to top-$100{\times}100$. This indicates that performance is more sensitive to the number of compositions during inference and signifies trade-off between performance and deployment constraints like storage and latency.

\begin{table}[ht]
\caption{Effect of the number of hard negatives for training (Train top-$k$), and the number of compositions selected for inference (Test top-$k_a {\times} k_o$) for unseen classes (mAP metric) on IG-504-144 dataset.
\vspace{-0.2cm}
}
\centering
\def\arraystretch{0.8} 
\def\cw{0.8cm}
\begin{tabular}{L{3cm}C{\cw}C{\cw}C{\cw}C{\cw}}
\toprule
\multirow{2}{*}{Test top-$k_a {\times} k_o$} & \multicolumn{4}{c}{Train top-$k$} \\
\cmidrule(lr){2-5}
& 900 & 2500 & 5000 & ALL \\
\midrule
$k_a=30$; $k_o=30$ & 9.70 & 9.82 & 9.84 & 9.84\\
$k_a=50$; $k_o=50$ & 10.15 & 10.27 & 10.28 & 10.29 \\
$k_a=100$; $k_o=100$ & 10.63 & 10.74 & 10.73 & 10.76 \\
\bottomrule
\end{tabular}
\label{tab:output_size}
\end{table}

\subsection{Comparison with baselines}

\paragraph{Baselines.}
First, let us describe two simple baselines that are trained on the same data as our approach.

\noindent{\it{Composition FC.}}
Instead of our composition head, we use a fully-connected (FC) layer to learn attribute-object classifiers directly. Note that only compositions observed at training can be learned using this approach. Hence, it is impossible to evaluate it on the unseen set. However, this method is still useful as an upper bound for composing classifiers on the seen attribute-object pairs. We only learn it on the smaller IG-504-144 dataset, where the number of compositions available at training is feasible.

\noindent{\it{Softmax Product.}}
In this baseline, the composition head is removed from the pipeline, and a late fusion of individual attribute and object scores is performed.
More precisely, joint attribute-object probability is estimated as a product of softmax probabilities from the attribute and object head of the system, \ie, $p_{ao}=p_a p_o$.

\paragraph{Instagram Datasets.}
We compare CompNet 
pipeline against the baselines on Instagram datasets. There is a small but consistent performance improvement on standard object and attribute tasks, see Table~\ref{tab:comparison_obj_attr}.
We attribute this to the fact that our composition head also benefits the image representation that is being trained jointly.

Table~\ref{tab:comparison} shows the performance of different approaches on both seen and unseen attribute-object pairs. Our approach shows significant improvement over the Softmax Product baseline. This is especially prominent in the unseen case, where mAP improvement is $+4.0$ and $+3.9$ on IG-504-144 and IG-8k-1k, respectively.
Additionally, as we increase the data scale, the gap between seen and unseen setup reduces significantly from $8.1$ mAP ($19.3$ (S),  $11.2$ (U)) to  $0.9$ mAP ($29.7$ (S),  $28.8$ (U)). In contrast, note that the gap is $2.7$ mAP for Softmax Product baseline compared to $0.9$ with our approach for IG-8k-1k dataset. Finally, directly training linear classifiers on compositions (Composition FC) is just slightly better ($+0.3$ mAP) than our approach. This additionally amplifies the benefit of using composition network, which also generalizes to attribute-object pairs with no training data.

\begin{table}[t]
\caption{
Performance evaluation via precision@1 of the object and attribute heads on Instagram datasets. 
\vspace{-0.2cm}
}
\centering
\def\arraystretch{0.8} 
\def\cw{1cm}
\begin{tabular}{L{2.5cm}C{\cw}C{\cw}C{\cw}C{\cw}}
\toprule
\multirow{2}{*}{Method} & \multicolumn{2}{c}{IG-504-144} &  \multicolumn{2}{c}{IG-8k-1k} \\
\cmidrule(lr){2-3} \cmidrule(lr){4-5} 
& Obj. & Attr. & Obj. & Attr.\\
\midrule
Composition FC & 70.9 & 29.9 & -- & -- \\
Softmax Product & 70.4 & 29.9 & 38.6 & 31.4 \\
\textbf{CompNet} & 71.0 & 30.0 & 38.9 & 31.7 \\
\bottomrule
\end{tabular}
\label{tab:comparison_obj_attr}
\end{table}

\begin{table}[t]
\caption{Performance evaluation of the composition head on Instagram datasets. Performance is reported as mAP averaged over seen (S) and unseen (U) composition splits.
\vspace{-0.1cm}
}
\centering
\def\arraystretch{0.8} 
\def\cw{0.7cm}
\begin{tabular}{L{3.2cm}C{\cw}C{\cw}C{\cw}C{\cw}}
\toprule
\multirow{2}{*}{Method} & \multicolumn{2}{c}{IG-504-144} &  \multicolumn{2}{c}{IG-8k-1k}\\
\cmidrule(lr){2-3} \cmidrule(lr){4-5} 
& S & U & S & U \\
\midrule
Composition FC & 19.6 & -- & -- & -- \\
Softmax Product & 18.2 & 7.2 & 27.6 & 24.9 \\
\textbf{CompNet} & 19.3 & 11.2 & 29.7 & 28.8 \\
\bottomrule
\end{tabular}
\label{tab:comparison}
\end{table}

\paragraph{Marketplace Datasets.}
Similar observations are made when evaluating pipeline on our internal Marketplace dataset, see Table~\ref{tab:marketplace}. We see significant improvements in both seen and unseen scenarios. This indicates the consistency of our evaluation and results between the weakly-supervised and fully-supervised settings.

\begin{table}[t]
\caption{Performance evaluation of the entire framework (object, attribute, and, composition head) on Marketplace dataset. Performance is reported as precision@1 for objects, mAP for attributes, and, mAP averaged over seen (S) and unseen (U) composition splits.
\vspace{-0.1cm}
}
\centering
\def\arraystretch{0.8} 
\def\cw{1.2cm}
\begin{tabular}{L{2cm}C{\cw}C{\cw}C{\cw}C{\cw}}
\toprule
Method & Object & Attribute & \multicolumn{2}{c}{Compositions} \\
& & & seen & unseen \\
\midrule
Softmax Prod. & 95.9 & 55.1 & 47.5 & 41.9 \\
\textbf{CompNet} & 96.0 & 56.4 & 50.1 & 46.9 \\
\bottomrule
\end{tabular}
\label{tab:marketplace}
\end{table}

\begin{figure}[t]
\centering
\input{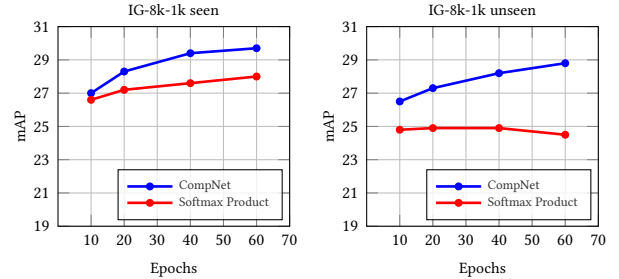}

\begin{tabular}{cc}

\begin{tikzpicture}
\begin{axis}[%
	ylabel near ticks, ylabel shift = -3 pt, yticklabel pos=left,
	xlabel near ticks,
	font=\scriptsize,
	width=0.55\linewidth,
	height=0.50\linewidth,
	xlabel={Epochs},
	ylabel={mAP},
	xlabel style  = {yshift = 0pt},
	title={IG-8k-1k seen},
	legend pos=south east,
    legend style={cells={anchor=west}, font =\tiny, fill opacity=0.8, row sep=-2.5pt},
    ymin = 19,
    ymax = 31,
    xmin = 0,
    xmax = 70,
    grid=both,
    xtick={10,20,...,100},
    ytick={1, 3, ..., 100},
    title style = {yshift = -5pt}
]  
    \addplot[color=blue, solid, mark=*, mark size=1, line width=1] table[x=epochs, y expr={\thisrow{compnetseen}}] \iglargeeps; \leg{CompNet};
    \addplot[color=red, solid, mark=*, mark size=1, line width=1] table[x=epochs, y expr={\thisrow{softprodseen}}] \iglargeeps; \leg{Softmax Product};
\end{axis}
\end{tikzpicture}

\hspace{-0.3cm}
&

\begin{tikzpicture}
\begin{axis}[%
	ylabel near ticks, ylabel shift = -3 pt, yticklabel pos=left,
	xlabel near ticks,
	font=\scriptsize,
	width=0.55\linewidth,
	height=0.50\linewidth,
	xlabel={Epochs},
	ylabel={mAP},
	xlabel style  = {yshift = 0pt},
	title={IG-8k-1k unseen},
	legend pos=south east,
    legend style={cells={anchor=west}, font =\tiny, fill opacity=0.8, row sep=-2.5pt},
    ymin = 19,
    ymax = 31,
    xmin = 0,
    xmax = 70,
    grid=both,
    xtick={10,20,...,100},
    ytick={1, 3, ..., 100},
    title style = {yshift = -5pt}
]  
    \addplot[color=blue, solid, mark=*, mark size=1, line width=1] table[x=epochs, y expr={\thisrow{compnetunseen}}] \iglargeeps; \leg{CompNet};
    \addplot[color=red, solid, mark=*, mark size=1, line width=1] table[x=epochs, y expr={\thisrow{softprodunseen}}] \iglargeeps; \leg{Softmax Product};
\end{axis}
\end{tikzpicture}

\end{tabular}
\vspace{-0.2cm}
\caption{Evaluating overfitting to seen compositions on Instagram datasets for varying lengths of the training.
\label{fig:epochs}
}
\end{figure}

\paragraph{Effect of Number of Training Epochs.}
Figure~\ref{fig:epochs} shows the results on seen and unseen compositions as we increase the number of training epochs for IG-8k-1k dataset. Note that, the plots do not present one training run evaluated at different stages of training. Rather, learning rate decay schedule is completed in full for each point on the plot. We observe that Softmax Product saturates or starts overfitting to seen compositions very early, \ie, after $20$ epochs. On the other hand, our pipeline continues improving unseen categories for a longer number of epochs, which in turn helps tighten the gap between seen and unseen performance.

\subsection{Qualitative results}
\begin{figure}
\input{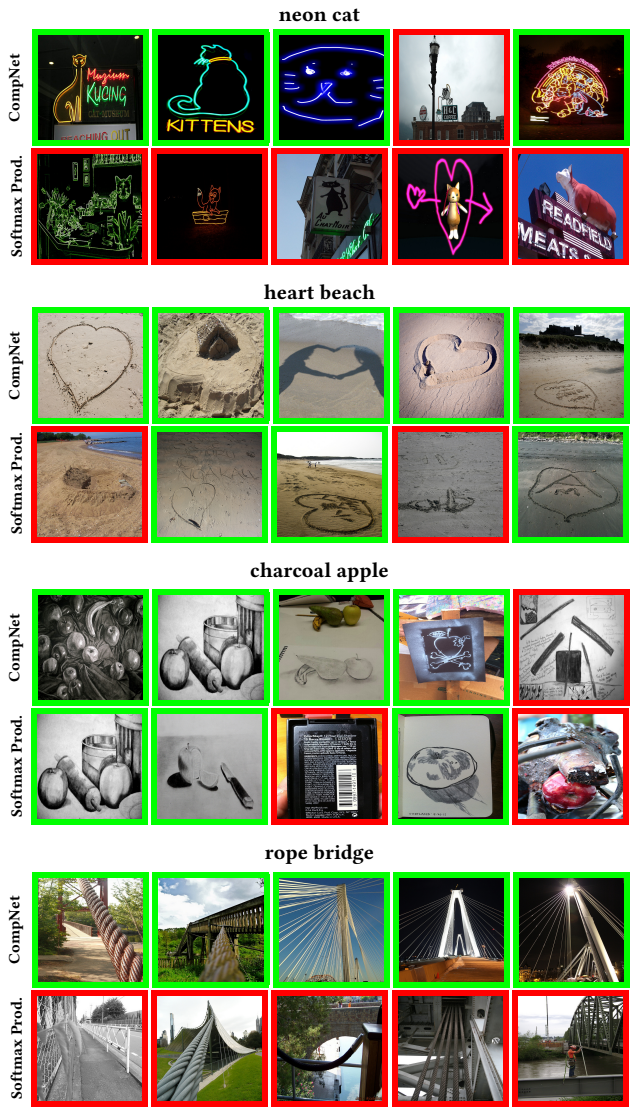}
\vspace{-0.1cm}
\caption{Qualitative comparison between our proposed CompNet (upper row) and Softmax Product (lower row) on \emph{unseen} compositions from images of YFCC100m. Every $10^{th}$ ranked image is shown to capture a larger ranking.
\label{fig:yfcc}
}
\end{figure}

We extract predictions on the YFCC100m dataset~\cite{TSF+15} due to large number of images and a high variation of fine-grained classes, and because we cannot show images from Instagram or Marketplace.
YFCC100m does not have composition annotations, so we randomly picked 50 \emph{unseen} compositions and inspected image shortlist with highest composition prediction scores. 
We compare our CompNet method with Softmax Product baseline, both trained on IG-8k-1k, and did not find any composition in which Softmax Product performed better.
Few example compositions are depicted in Figure~\ref{fig:yfcc}.

%


\section{Deployment}
\label{sec:deployment}
\emph{CompNet} hasn't been deployed at Facebook yet, but we have identified the first few large scale use-cases within Facebook Marketplace and have run experiments on the Marketplace data to verify the effectiveness of this approach for applications in Commerce.

\subsection{Sample Use Cases}
\paragraph{Marketplace Attributes Filters.}
Users on Marketplace have access to filters which enable them to search items for the presence of a specific attribute within a category page. For instance, searching for products with the \emph{red color} within the \emph{Women's Dresses} category. A major limitation to the currently deployed system is that the model predicts category specific attributes (\emph{red dress, peplum sweater, etc}), each being learnt as a linear classifier, thus restricting the attributes to certain handpicked categories. So, adding a new category requires getting annotations for each applicable attribute. With \emph{CompNet}, we can extend the set of categories to thousands without requiring new annotations because the attributes head is category agnostic. Hence, at inference time we could compose new attributes for unseen categories for free. This will enable us in increasing the coverage of attributes over the total inventory and  adding new attributes filters which were not possible before.

\paragraph{Marketplace Browse Feed Ranking.}
Browse Feed Ranking is a system that ranks the products on the homepage for all Marketplace users. The current ranking system uses a variety of features from multiple sources (\emph{user, image, and text}). Our hypothesis is that having features related to fine-grained attributes on an image will help rank products better for every individual user. However, with handpicked categories, coverage is again an issue. 
With \emph{CompNet}, we will be able to predict attributes over multiple categories (\emph{seen and unseen}), thus improving the overall coverage, which might lead to downstream gains in ranking.

\subsection{Deployment Challenges \& Plan}
There are some challenges in deploying the \emph{CompNet} system in production from a compute and storage point of view. We decide not to use the MLP during inference and directly deploy the linear compositional classifiers so that the inference system is agnostic to the type of head used to construct the classifiers and simplify future upgrades. Assuming that \emph{CompNet} predicts 1000 categories and 1000 attributes during inference, we have around 1M classifiers for the compositions. But, we do not need to deploy all 1M classifiers as from our past knowledge, we know that some attributes never occur with some categories (\eg \emph{neckline} attributes never occur with \emph{Home \& Garden} categories). So, we can significantly reduce the number of classifiers that we need to store at all times (in the order of few hundred of thousands or potentially even less). Hence, based on the inference strategy mentioned in Section~\ref{sec:inference}, we load the top $k_a \times k_o$ classifiers out of the few hundreds of thousands deployed classifiers and compute scores.

After \emph{CompNet} predictions are available during inference, we'll store the outputs in a distributed key-value store, which would be consumed by multiple product groups. 
We go a step further and store only the top few (\textasciitilde 100) compositional scores instead of $k_a \times k_o$ in permanent storage for consumption.


\section{Conclusion and Future Work}
\label{sec:conclusion}

In this work we explored a framework for joint attribute-object composition classification, learned from a large-scale weakly-supervised dataset, combining 8k objects and 1k attributes.
Carefully designed loss functions enable us to handle noisy labels and generalize to compositions not present in the training data. Extensive evaluations show the benefits of using such an approach compared to a late fusion of individual attribute and object predictions.

There are few challenges we leave for future research.
Namely, for a given image, training compositions are selected as every pairwise combination of attribute and object hashtags.
When there is more than one attribute and/or more than one object, combining all of them will likely add noise. 
We plan on exploring weak localization techniques to alleviate this problem.
Additionally, some attributes commonly occur with a specific object, and sparsely with others.
For example, \texttt{\#striped} always occurs with \texttt{\#zebra}, and sparsely with \texttt{\#shirt}, \texttt{\#wall}, \texttt{\#couch}, thus being biased towards a specific object category.
We have not quantified such bias, but are planning to do so in an attempt to develop methods that reduce it.



\newpage
\bibliographystyle{ACM-Reference-Format}
\bibliography{bib}


\begin{thebibliography}{62}


\ifx \showCODEN    \undefined \def \showCODEN     #1{\unskip}     \fi
\ifx \showDOI      \undefined \def \showDOI       #1{#1}\fi
\ifx \showISBNx    \undefined \def \showISBNx     #1{\unskip}     \fi
\ifx \showISBNxiii \undefined \def \showISBNxiii  #1{\unskip}     \fi
\ifx \showISSN     \undefined \def \showISSN      #1{\unskip}     \fi
\ifx \showLCCN     \undefined \def \showLCCN      #1{\unskip}     \fi
\ifx \shownote     \undefined \def \shownote      #1{#1}          \fi
\ifx \showarticletitle \undefined \def \showarticletitle #1{#1}   \fi
\ifx \showURL      \undefined \def \showURL       {\relax}        \fi
\providecommand\bibfield[2]{#2}
\providecommand\bibinfo[2]{#2}
\providecommand\natexlab[1]{#1}
\providecommand\showeprint[2][]{arXiv:#2}

\bibitem[\protect\citeauthoryear{Al-Halah, Tapaswi, and Stiefelhagen}{Al-Halah
  et~al\mbox{.}}{2016}]%
        {ATS16}
\bibfield{author}{\bibinfo{person}{Ziad Al-Halah}, \bibinfo{person}{Makarand
  Tapaswi}, {and} \bibinfo{person}{Rainer Stiefelhagen}.}
  \bibinfo{year}{2016}\natexlab{}.
\newblock \showarticletitle{Recovering the missing link: Predicting
  class-attribute associations for unsupervised zero-shot learning}. In
  \bibinfo{booktitle}{\emph{CVPR}}.
\newblock


\bibitem[\protect\citeauthoryear{Baroni and Zamparelli}{Baroni and
  Zamparelli}{2010}]%
        {BZ10}
\bibfield{author}{\bibinfo{person}{Marco Baroni} {and} \bibinfo{person}{Roberto
  Zamparelli}.} \bibinfo{year}{2010}\natexlab{}.
\newblock \showarticletitle{Nouns are vectors, adjectives are matrices:
  Representing adjective-noun constructions in semantic space}. In
  \bibinfo{booktitle}{\emph{EMNLP}}.
\newblock


\bibitem[\protect\citeauthoryear{Berg, Berg, and Shih}{Berg
  et~al\mbox{.}}{2010}]%
        {BBS10}
\bibfield{author}{\bibinfo{person}{Tamara~L Berg}, \bibinfo{person}{Alexander~C
  Berg}, {and} \bibinfo{person}{Jonathan Shih}.}
  \bibinfo{year}{2010}\natexlab{}.
\newblock \showarticletitle{Automatic attribute discovery and characterization
  from noisy web data}. In \bibinfo{booktitle}{\emph{ECCV}}.
\newblock


\bibitem[\protect\citeauthoryear{Chen and Grauman}{Chen and Grauman}{2014}]%
        {CG14}
\bibfield{author}{\bibinfo{person}{Chao-Yeh Chen} {and}
  \bibinfo{person}{Kristen Grauman}.} \bibinfo{year}{2014}\natexlab{}.
\newblock \showarticletitle{Inferring analogous attributes}. In
  \bibinfo{booktitle}{\emph{CVPR}}.
\newblock


\bibitem[\protect\citeauthoryear{Chen, Gallagher, and Girod}{Chen
  et~al\mbox{.}}{2012}]%
        {CGG12}
\bibfield{author}{\bibinfo{person}{Huizhong Chen}, \bibinfo{person}{Andrew
  Gallagher}, {and} \bibinfo{person}{Bernd Girod}.}
  \bibinfo{year}{2012}\natexlab{}.
\newblock \showarticletitle{Describing clothing by semantic attributes}. In
  \bibinfo{booktitle}{\emph{ECCV}}.
\newblock


\bibitem[\protect\citeauthoryear{Chen, Nan, Jingjing, and Zheng}{Chen
  et~al\mbox{.}}{2020}]%
        {CNJZ20}
\bibfield{author}{\bibinfo{person}{Hui Chen}, \bibinfo{person}{Zhixiong Nan},
  \bibinfo{person}{Jiang Jingjing}, {and} \bibinfo{person}{Nanning Zheng}.}
  \bibinfo{year}{2020}\natexlab{}.
\newblock \showarticletitle{Learning to Infer Unseen Attribute-Object
  Compositions}.
\newblock \bibinfo{journal}{\emph{arXiv:2010.14343}} (\bibinfo{year}{2020}).
\newblock


\bibitem[\protect\citeauthoryear{Deng, Luo, Loy, and Tang}{Deng
  et~al\mbox{.}}{2014}]%
        {DLLT14}
\bibfield{author}{\bibinfo{person}{Yubin Deng}, \bibinfo{person}{Ping Luo},
  \bibinfo{person}{Chen~Change Loy}, {and} \bibinfo{person}{Xiaoou Tang}.}
  \bibinfo{year}{2014}\natexlab{}.
\newblock \showarticletitle{Pedestrian attribute recognition at far distance}.
  In \bibinfo{booktitle}{\emph{ACM-MM}}.
\newblock


\bibitem[\protect\citeauthoryear{Denton, Weston, Paluri, Bourdev, and
  Fergus}{Denton et~al\mbox{.}}{2015}]%
        {DWP+15}
\bibfield{author}{\bibinfo{person}{Emily Denton}, \bibinfo{person}{Jason
  Weston}, \bibinfo{person}{Manohar Paluri}, \bibinfo{person}{Lubomir Bourdev},
  {and} \bibinfo{person}{Rob Fergus}.} \bibinfo{year}{2015}\natexlab{}.
\newblock \showarticletitle{User conditional hashtag prediction for images}. In
  \bibinfo{booktitle}{\emph{SIGKDD}}.
\newblock


\bibitem[\protect\citeauthoryear{Duan, Parikh, Crandall, and Grauman}{Duan
  et~al\mbox{.}}{2012}]%
        {DPCG12}
\bibfield{author}{\bibinfo{person}{Kun Duan}, \bibinfo{person}{Devi Parikh},
  \bibinfo{person}{David Crandall}, {and} \bibinfo{person}{Kristen Grauman}.}
  \bibinfo{year}{2012}\natexlab{}.
\newblock \showarticletitle{Discovering localized attributes for fine-grained
  recognition}. In \bibinfo{booktitle}{\emph{CVPR}}.
\newblock


\bibitem[\protect\citeauthoryear{Farhadi, Endres, Hoiem, and Forsyth}{Farhadi
  et~al\mbox{.}}{2009}]%
        {FEHF09}
\bibfield{author}{\bibinfo{person}{Ali Farhadi}, \bibinfo{person}{Ian Endres},
  \bibinfo{person}{Derek Hoiem}, {and} \bibinfo{person}{David Forsyth}.}
  \bibinfo{year}{2009}\natexlab{}.
\newblock \showarticletitle{Describing objects by their attributes}. In
  \bibinfo{booktitle}{\emph{CVPR}}.
\newblock


\bibitem[\protect\citeauthoryear{Georgopoulos, Chrysos, Pantic, and
  Panagakis}{Georgopoulos et~al\mbox{.}}{2020}]%
        {GCPP20}
\bibfield{author}{\bibinfo{person}{Markos Georgopoulos},
  \bibinfo{person}{Grigorios Chrysos}, \bibinfo{person}{Maja Pantic}, {and}
  \bibinfo{person}{Yannis Panagakis}.} \bibinfo{year}{2020}\natexlab{}.
\newblock \showarticletitle{Multilinear Latent Conditioning for Generating
  Unseen Attribute Combinations}. In \bibinfo{booktitle}{\emph{ICML}}.
\newblock


\bibitem[\protect\citeauthoryear{Goodfellow, Bengio, and Courville}{Goodfellow
  et~al\mbox{.}}{2016}]%
        {GBC16}
\bibfield{author}{\bibinfo{person}{Ian Goodfellow}, \bibinfo{person}{Yoshua
  Bengio}, {and} \bibinfo{person}{Aaron Courville}.}
  \bibinfo{year}{2016}\natexlab{}.
\newblock \bibinfo{booktitle}{\emph{Deep learning}}.
\newblock \bibinfo{publisher}{MIT press Cambridge}.
\newblock


\bibitem[\protect\citeauthoryear{Goyal, Doll{\'a}r, Girshick, Noordhuis,
  Wesolowski, Kyrola, Tulloch, Jia, and He}{Goyal et~al\mbox{.}}{2017}]%
        {GDG+17}
\bibfield{author}{\bibinfo{person}{Priya Goyal}, \bibinfo{person}{Piotr
  Doll{\'a}r}, \bibinfo{person}{Ross Girshick}, \bibinfo{person}{Pieter
  Noordhuis}, \bibinfo{person}{Lukasz Wesolowski}, \bibinfo{person}{Aapo
  Kyrola}, \bibinfo{person}{Andrew Tulloch}, \bibinfo{person}{Yangqing Jia},
  {and} \bibinfo{person}{Kaiming He}.} \bibinfo{year}{2017}\natexlab{}.
\newblock \showarticletitle{Accurate, large minibatch sgd: Training imagenet in
  1 hour}.
\newblock \bibinfo{journal}{\emph{arXiv:1706.02677}} (\bibinfo{year}{2017}).
\newblock


\bibitem[\protect\citeauthoryear{Gross, Ranzato, and Szlam}{Gross
  et~al\mbox{.}}{2017}]%
        {GRS17}
\bibfield{author}{\bibinfo{person}{Sam Gross}, \bibinfo{person}{Marc'Aurelio
  Ranzato}, {and} \bibinfo{person}{Arthur Szlam}.}
  \bibinfo{year}{2017}\natexlab{}.
\newblock \showarticletitle{Hard mixtures of experts for large scale weakly
  supervised vision}. In \bibinfo{booktitle}{\emph{CVPR}}.
\newblock


\bibitem[\protect\citeauthoryear{Guevara}{Guevara}{2010}]%
        {G10}
\bibfield{author}{\bibinfo{person}{Emiliano~Raul Guevara}.}
  \bibinfo{year}{2010}\natexlab{}.
\newblock \showarticletitle{A regression model of adjective-noun
  compositionality in distributional semantics}. In
  \bibinfo{booktitle}{\emph{GEMS}}.
\newblock


\bibitem[\protect\citeauthoryear{He, Zhang, Ren, and Sun}{He
  et~al\mbox{.}}{2016}]%
        {HZRS16}
\bibfield{author}{\bibinfo{person}{Kaiming He}, \bibinfo{person}{Xiangyu
  Zhang}, \bibinfo{person}{Shaoqing Ren}, {and} \bibinfo{person}{Jian Sun}.}
  \bibinfo{year}{2016}\natexlab{}.
\newblock \showarticletitle{Deep residual learning for image recognition}. In
  \bibinfo{booktitle}{\emph{CVPR}}.
\newblock


\bibitem[\protect\citeauthoryear{Hsiao and Grauman}{Hsiao and Grauman}{2017}]%
        {HG17}
\bibfield{author}{\bibinfo{person}{Wei-Lin Hsiao} {and}
  \bibinfo{person}{Kristen Grauman}.} \bibinfo{year}{2017}\natexlab{}.
\newblock \showarticletitle{Learning the latent "look": Unsupervised discovery
  of a style-coherent embedding from fashion images}. In
  \bibinfo{booktitle}{\emph{ICCV}}.
\newblock


\bibitem[\protect\citeauthoryear{Ioffe and Szegedy}{Ioffe and Szegedy}{2015}]%
        {IS15}
\bibfield{author}{\bibinfo{person}{Sergey Ioffe} {and}
  \bibinfo{person}{Christian Szegedy}.} \bibinfo{year}{2015}\natexlab{}.
\newblock \showarticletitle{Batch normalization: Accelerating deep network
  training by reducing internal covariate shift}.
\newblock \bibinfo{journal}{\emph{arXiv:1502.03167}} (\bibinfo{year}{2015}).
\newblock


\bibitem[\protect\citeauthoryear{Isola, Lim, and Adelson}{Isola
  et~al\mbox{.}}{2015}]%
        {ILA15}
\bibfield{author}{\bibinfo{person}{Phillip Isola}, \bibinfo{person}{Joseph~J
  Lim}, {and} \bibinfo{person}{Edward~H Adelson}.}
  \bibinfo{year}{2015}\natexlab{}.
\newblock \showarticletitle{Discovering states and transformations in image
  collections}. In \bibinfo{booktitle}{\emph{CVPR}}.
\newblock


\bibitem[\protect\citeauthoryear{Joulin, Van Der~Maaten, Jabri, and
  Vasilache}{Joulin et~al\mbox{.}}{2016}]%
        {JVJV16}
\bibfield{author}{\bibinfo{person}{Armand Joulin}, \bibinfo{person}{Laurens Van
  Der~Maaten}, \bibinfo{person}{Allan Jabri}, {and} \bibinfo{person}{Nicolas
  Vasilache}.} \bibinfo{year}{2016}\natexlab{}.
\newblock \showarticletitle{Learning visual features from large weakly
  supervised data}. In \bibinfo{booktitle}{\emph{ECCV}}.
\newblock


\bibitem[\protect\citeauthoryear{Kovashka, Parikh, and Grauman}{Kovashka
  et~al\mbox{.}}{2012}]%
        {KPG12}
\bibfield{author}{\bibinfo{person}{Adriana Kovashka}, \bibinfo{person}{Devi
  Parikh}, {and} \bibinfo{person}{Kristen Grauman}.}
  \bibinfo{year}{2012}\natexlab{}.
\newblock \showarticletitle{Whittlesearch: Image search with relative attribute
  feedback}. In \bibinfo{booktitle}{\emph{CVPR}}.
\newblock


\bibitem[\protect\citeauthoryear{Krizhevsky, Sutskever, and Hinton}{Krizhevsky
  et~al\mbox{.}}{2012}]%
        {KSH12}
\bibfield{author}{\bibinfo{person}{Alex Krizhevsky}, \bibinfo{person}{Ilya
  Sutskever}, {and} \bibinfo{person}{Geoffrey~E Hinton}.}
  \bibinfo{year}{2012}\natexlab{}.
\newblock \showarticletitle{Imagenet classification with deep convolutional
  neural networks}. In \bibinfo{booktitle}{\emph{NeurIPS}}.
\newblock


\bibitem[\protect\citeauthoryear{Kumar, Belhumeur, and Nayar}{Kumar
  et~al\mbox{.}}{2008}]%
        {KBN08}
\bibfield{author}{\bibinfo{person}{Neeraj Kumar}, \bibinfo{person}{Peter
  Belhumeur}, {and} \bibinfo{person}{Shree Nayar}.}
  \bibinfo{year}{2008}\natexlab{}.
\newblock \showarticletitle{Facetracer: A search engine for large collections
  of images with faces}. In \bibinfo{booktitle}{\emph{ECCV}}.
\newblock


\bibitem[\protect\citeauthoryear{Kumar, Berg, Belhumeur, and Nayar}{Kumar
  et~al\mbox{.}}{2011}]%
        {KBBN11}
\bibfield{author}{\bibinfo{person}{Neeraj Kumar}, \bibinfo{person}{Alexander
  Berg}, \bibinfo{person}{Peter~N Belhumeur}, {and} \bibinfo{person}{Shree
  Nayar}.} \bibinfo{year}{2011}\natexlab{}.
\newblock \showarticletitle{Describable visual attributes for face verification
  and image search}.
\newblock \bibinfo{journal}{\emph{TPAMI}} (\bibinfo{year}{2011}).
\newblock


\bibitem[\protect\citeauthoryear{Kumar, Berg, Belhumeur, and Nayar}{Kumar
  et~al\mbox{.}}{2009}]%
        {KBBN09}
\bibfield{author}{\bibinfo{person}{Neeraj Kumar}, \bibinfo{person}{Alexander~C
  Berg}, \bibinfo{person}{Peter~N Belhumeur}, {and} \bibinfo{person}{Shree~K
  Nayar}.} \bibinfo{year}{2009}\natexlab{}.
\newblock \showarticletitle{Attribute and simile classifiers for face
  verification}. In \bibinfo{booktitle}{\emph{ICCV}}.
\newblock


\bibitem[\protect\citeauthoryear{Laffont, Ren, Tao, Qian, and Hays}{Laffont
  et~al\mbox{.}}{2014}]%
        {LRT+14}
\bibfield{author}{\bibinfo{person}{Pierre-Yves Laffont}, \bibinfo{person}{Zhile
  Ren}, \bibinfo{person}{Xiaofeng Tao}, \bibinfo{person}{Chao Qian}, {and}
  \bibinfo{person}{James Hays}.} \bibinfo{year}{2014}\natexlab{}.
\newblock \showarticletitle{Transient attributes for high-level understanding
  and editing of outdoor scenes}.
\newblock \bibinfo{journal}{\emph{TOG}} (\bibinfo{year}{2014}).
\newblock


\bibitem[\protect\citeauthoryear{Lampert, Nickisch, and Harmeling}{Lampert
  et~al\mbox{.}}{2009}]%
        {LNH09}
\bibfield{author}{\bibinfo{person}{Christoph~H Lampert},
  \bibinfo{person}{Hannes Nickisch}, {and} \bibinfo{person}{Stefan Harmeling}.}
  \bibinfo{year}{2009}\natexlab{}.
\newblock \showarticletitle{Learning to detect unseen object classes by
  between-class attribute transfer}. In \bibinfo{booktitle}{\emph{CVPR}}.
\newblock


\bibitem[\protect\citeauthoryear{Li, Jabri, Joulin, and van~der Maaten}{Li
  et~al\mbox{.}}{2017}]%
        {LJJV17}
\bibfield{author}{\bibinfo{person}{Ang Li}, \bibinfo{person}{Allan Jabri},
  \bibinfo{person}{Armand Joulin}, {and} \bibinfo{person}{Laurens van~der
  Maaten}.} \bibinfo{year}{2017}\natexlab{}.
\newblock \showarticletitle{Learning visual n-grams from web data}. In
  \bibinfo{booktitle}{\emph{ICCV}}.
\newblock


\bibitem[\protect\citeauthoryear{Liu, Kuipers, and Savarese}{Liu
  et~al\mbox{.}}{2011}]%
        {LKS11}
\bibfield{author}{\bibinfo{person}{Jingen Liu}, \bibinfo{person}{Benjamin
  Kuipers}, {and} \bibinfo{person}{Silvio Savarese}.}
  \bibinfo{year}{2011}\natexlab{}.
\newblock \showarticletitle{Recognizing human actions by attributes}. In
  \bibinfo{booktitle}{\emph{CVPR}}.
\newblock


\bibitem[\protect\citeauthoryear{Liu, Luo, Qiu, Wang, and Tang}{Liu
  et~al\mbox{.}}{2016}]%
        {LLQ+16}
\bibfield{author}{\bibinfo{person}{Ziwei Liu}, \bibinfo{person}{Ping Luo},
  \bibinfo{person}{Shi Qiu}, \bibinfo{person}{Xiaogang Wang}, {and}
  \bibinfo{person}{Xiaoou Tang}.} \bibinfo{year}{2016}\natexlab{}.
\newblock \showarticletitle{DeepFashion: Powering Robust Clothes Recognition
  and Retrieval with Rich Annotations}. In \bibinfo{booktitle}{\emph{CVPR}}.
\newblock


\bibitem[\protect\citeauthoryear{Liu, Luo, Wang, and Tang}{Liu
  et~al\mbox{.}}{2015}]%
        {LLWT15}
\bibfield{author}{\bibinfo{person}{Ziwei Liu}, \bibinfo{person}{Ping Luo},
  \bibinfo{person}{Xiaogang Wang}, {and} \bibinfo{person}{Xiaoou Tang}.}
  \bibinfo{year}{2015}\natexlab{}.
\newblock \showarticletitle{Deep learning face attributes in the wild}. In
  \bibinfo{booktitle}{\emph{ICCV}}.
\newblock


\bibitem[\protect\citeauthoryear{Loshchilov and Hutter}{Loshchilov and
  Hutter}{2016}]%
        {LH16}
\bibfield{author}{\bibinfo{person}{Ilya Loshchilov} {and}
  \bibinfo{person}{Frank Hutter}.} \bibinfo{year}{2016}\natexlab{}.
\newblock \showarticletitle{Sgdr: Stochastic gradient descent with warm
  restarts}.
\newblock \bibinfo{journal}{\emph{arXiv:1608.03983}} (\bibinfo{year}{2016}).
\newblock


\bibitem[\protect\citeauthoryear{Lu, Kumar, Zhai, Cheng, Javidi, and Feris}{Lu
  et~al\mbox{.}}{2017}]%
        {LKZ+17}
\bibfield{author}{\bibinfo{person}{Yongxi Lu}, \bibinfo{person}{Abhishek
  Kumar}, \bibinfo{person}{Shuangfei Zhai}, \bibinfo{person}{Yu Cheng},
  \bibinfo{person}{Tara Javidi}, {and} \bibinfo{person}{Rogerio Feris}.}
  \bibinfo{year}{2017}\natexlab{}.
\newblock \showarticletitle{Fully-adaptive feature sharing in multi-task
  networks with applications in person attribute classification}. In
  \bibinfo{booktitle}{\emph{CVPR}}.
\newblock


\bibitem[\protect\citeauthoryear{Maas, Hannun, and Ng}{Maas
  et~al\mbox{.}}{2013}]%
        {MHN13}
\bibfield{author}{\bibinfo{person}{Andrew~L Maas}, \bibinfo{person}{Awni~Y
  Hannun}, {and} \bibinfo{person}{Andrew~Y Ng}.}
  \bibinfo{year}{2013}\natexlab{}.
\newblock \showarticletitle{Rectifier nonlinearities improve neural network
  acoustic models}. In \bibinfo{booktitle}{\emph{ICML}}.
\newblock


\bibitem[\protect\citeauthoryear{Mahajan, Girshick, Ramanathan, He, Paluri, Li,
  Bharambe, and van~der Maaten}{Mahajan et~al\mbox{.}}{2018}]%
        {MGR+18}
\bibfield{author}{\bibinfo{person}{Dhruv Mahajan}, \bibinfo{person}{Ross
  Girshick}, \bibinfo{person}{Vignesh Ramanathan}, \bibinfo{person}{Kaiming
  He}, \bibinfo{person}{Manohar Paluri}, \bibinfo{person}{Yixuan Li},
  \bibinfo{person}{Ashwin Bharambe}, {and} \bibinfo{person}{Laurens van~der
  Maaten}.} \bibinfo{year}{2018}\natexlab{}.
\newblock \showarticletitle{Exploring the limits of weakly supervised
  pretraining}. In \bibinfo{booktitle}{\emph{ECCV}}.
\newblock


\bibitem[\protect\citeauthoryear{Miller}{Miller}{1995}]%
        {M95}
\bibfield{author}{\bibinfo{person}{George~A Miller}.}
  \bibinfo{year}{1995}\natexlab{}.
\newblock \showarticletitle{WordNet: a lexical database for English}.
\newblock \bibinfo{journal}{\emph{Commun. ACM}} (\bibinfo{year}{1995}).
\newblock


\bibitem[\protect\citeauthoryear{Misra, Gupta, and Hebert}{Misra
  et~al\mbox{.}}{2017}]%
        {MGH17}
\bibfield{author}{\bibinfo{person}{Ishan Misra}, \bibinfo{person}{Abhinav
  Gupta}, {and} \bibinfo{person}{Martial Hebert}.}
  \bibinfo{year}{2017}\natexlab{}.
\newblock \showarticletitle{From red wine to red tomato: Composition with
  context}. In \bibinfo{booktitle}{\emph{CVPR}}.
\newblock


\bibitem[\protect\citeauthoryear{Mitchell and Lapata}{Mitchell and
  Lapata}{2008}]%
        {ML08}
\bibfield{author}{\bibinfo{person}{Jeff Mitchell} {and}
  \bibinfo{person}{Mirella Lapata}.} \bibinfo{year}{2008}\natexlab{}.
\newblock \showarticletitle{Vector-based models of semantic composition}. In
  \bibinfo{booktitle}{\emph{ACL-HLT}}.
\newblock


\bibitem[\protect\citeauthoryear{Nagarajan and Grauman}{Nagarajan and
  Grauman}{2018}]%
        {NG18}
\bibfield{author}{\bibinfo{person}{Tushar Nagarajan} {and}
  \bibinfo{person}{Kristen Grauman}.} \bibinfo{year}{2018}\natexlab{}.
\newblock \showarticletitle{Attributes as operators: factorizing unseen
  attribute-object compositions}. In \bibinfo{booktitle}{\emph{ECCV}}.
\newblock


\bibitem[\protect\citeauthoryear{Nguyen, Lazaridou, and Bernardi}{Nguyen
  et~al\mbox{.}}{2014}]%
        {NLB14}
\bibfield{author}{\bibinfo{person}{Dat~Tien Nguyen}, \bibinfo{person}{Angeliki
  Lazaridou}, {and} \bibinfo{person}{Raffaella Bernardi}.}
  \bibinfo{year}{2014}\natexlab{}.
\newblock \showarticletitle{Coloring objects: adjective-noun visual semantic
  compositionality}. In \bibinfo{booktitle}{\emph{ACL Workshop on Vision and
  Language}}.
\newblock


\bibitem[\protect\citeauthoryear{Patterson and Hays}{Patterson and
  Hays}{2012}]%
        {PH12}
\bibfield{author}{\bibinfo{person}{Genevieve Patterson} {and}
  \bibinfo{person}{James Hays}.} \bibinfo{year}{2012}\natexlab{}.
\newblock \showarticletitle{Sun attribute database: Discovering, annotating,
  and recognizing scene attributes}. In \bibinfo{booktitle}{\emph{CVPR}}.
\newblock


\bibitem[\protect\citeauthoryear{Patterson and Hays}{Patterson and
  Hays}{2016}]%
        {PH16}
\bibfield{author}{\bibinfo{person}{Genevieve Patterson} {and}
  \bibinfo{person}{James Hays}.} \bibinfo{year}{2016}\natexlab{}.
\newblock \showarticletitle{Coco attributes: Attributes for people, animals,
  and objects}. In \bibinfo{booktitle}{\emph{ECCV}}.
\newblock


\bibitem[\protect\citeauthoryear{Pezzelle, Shekhar, and Bernardi}{Pezzelle
  et~al\mbox{.}}{2016}]%
        {PSB16}
\bibfield{author}{\bibinfo{person}{Sandro Pezzelle}, \bibinfo{person}{Ravi
  Shekhar}, {and} \bibinfo{person}{Raffaella Bernardi}.}
  \bibinfo{year}{2016}\natexlab{}.
\newblock \showarticletitle{Building a bagpipe with a bag and a pipe: Exploring
  conceptual combination in vision}. In \bibinfo{booktitle}{\emph{ACL Workshop
  on Vision and Language}}.
\newblock


\bibitem[\protect\citeauthoryear{Russakovsky, Deng, Su, Krause, Satheesh, Ma,
  Huang, Karpathy, Khosla, Bernstein, et~al\mbox{.}}{Russakovsky
  et~al\mbox{.}}{2015}]%
        {RDS+15}
\bibfield{author}{\bibinfo{person}{Olga Russakovsky}, \bibinfo{person}{Jia
  Deng}, \bibinfo{person}{Hao Su}, \bibinfo{person}{Jonathan Krause},
  \bibinfo{person}{Sanjeev Satheesh}, \bibinfo{person}{Sean Ma},
  \bibinfo{person}{Zhiheng Huang}, \bibinfo{person}{Andrej Karpathy},
  \bibinfo{person}{Aditya Khosla}, \bibinfo{person}{Michael Bernstein},
  {et~al\mbox{.}}} \bibinfo{year}{2015}\natexlab{}.
\newblock \showarticletitle{Imagenet large scale visual recognition challenge}.
\newblock \bibinfo{journal}{\emph{IJCV}} (\bibinfo{year}{2015}).
\newblock


\bibitem[\protect\citeauthoryear{Russakovsky and Fei-Fei}{Russakovsky and
  Fei-Fei}{2010}]%
        {RF10}
\bibfield{author}{\bibinfo{person}{Olga Russakovsky} {and} \bibinfo{person}{Li
  Fei-Fei}.} \bibinfo{year}{2010}\natexlab{}.
\newblock \showarticletitle{Attribute learning in large-scale datasets}. In
  \bibinfo{booktitle}{\emph{ECCV}}.
\newblock


\bibitem[\protect\citeauthoryear{Sadeghi and Farhadi}{Sadeghi and
  Farhadi}{2011}]%
        {SF11}
\bibfield{author}{\bibinfo{person}{Mohammad~Amin Sadeghi} {and}
  \bibinfo{person}{Ali Farhadi}.} \bibinfo{year}{2011}\natexlab{}.
\newblock \showarticletitle{Recognition using visual phrases}. In
  \bibinfo{booktitle}{\emph{CVPR}}.
\newblock


\bibitem[\protect\citeauthoryear{Santa~Cruz, Fernando, Cherian, and
  Gould}{Santa~Cruz et~al\mbox{.}}{2018}]%
        {SFCG18}
\bibfield{author}{\bibinfo{person}{Rodrigo Santa~Cruz}, \bibinfo{person}{Basura
  Fernando}, \bibinfo{person}{Anoop Cherian}, {and} \bibinfo{person}{Stephen
  Gould}.} \bibinfo{year}{2018}\natexlab{}.
\newblock \showarticletitle{Neural algebra of classifiers}. In
  \bibinfo{booktitle}{\emph{WACV}}.
\newblock


\bibitem[\protect\citeauthoryear{Sharma and Jurie}{Sharma and Jurie}{2011}]%
        {SJ11}
\bibfield{author}{\bibinfo{person}{Gaurav Sharma} {and}
  \bibinfo{person}{Frederic Jurie}.} \bibinfo{year}{2011}\natexlab{}.
\newblock \showarticletitle{Learning discriminative spatial representation for
  image classification}. In \bibinfo{booktitle}{\emph{BMVC}}.
\newblock


\bibitem[\protect\citeauthoryear{Singh and Lee}{Singh and Lee}{2016}]%
        {SL16}
\bibfield{author}{\bibinfo{person}{Krishna~Kumar Singh} {and}
  \bibinfo{person}{Yong~Jae Lee}.} \bibinfo{year}{2016}\natexlab{}.
\newblock \showarticletitle{End-to-end localization and ranking for relative
  attributes}. In \bibinfo{booktitle}{\emph{ECCV}}.
\newblock


\bibitem[\protect\citeauthoryear{Socher, Perelygin, Wu, Chuang, Manning, Ng,
  and Potts}{Socher et~al\mbox{.}}{2013}]%
        {SPW+13}
\bibfield{author}{\bibinfo{person}{Richard Socher}, \bibinfo{person}{Alex
  Perelygin}, \bibinfo{person}{Jean Wu}, \bibinfo{person}{Jason Chuang},
  \bibinfo{person}{Christopher~D Manning}, \bibinfo{person}{Andrew~Y Ng}, {and}
  \bibinfo{person}{Christopher Potts}.} \bibinfo{year}{2013}\natexlab{}.
\newblock \showarticletitle{Recursive deep models for semantic compositionality
  over a sentiment treebank}. In \bibinfo{booktitle}{\emph{EMNLP}}.
\newblock


\bibitem[\protect\citeauthoryear{Srivastava, Hinton, Krizhevsky, Sutskever, and
  Salakhutdinov}{Srivastava et~al\mbox{.}}{2014}]%
        {SHK+14}
\bibfield{author}{\bibinfo{person}{Nitish Srivastava},
  \bibinfo{person}{Geoffrey Hinton}, \bibinfo{person}{Alex Krizhevsky},
  \bibinfo{person}{Ilya Sutskever}, {and} \bibinfo{person}{Ruslan
  Salakhutdinov}.} \bibinfo{year}{2014}\natexlab{}.
\newblock \showarticletitle{Dropout: a simple way to prevent neural networks
  from overfitting}.
\newblock \bibinfo{journal}{\emph{JMLR}} (\bibinfo{year}{2014}).
\newblock


\bibitem[\protect\citeauthoryear{Su, Zhang, Xing, Gao, and Tian}{Su
  et~al\mbox{.}}{2016}]%
        {SZX+16}
\bibfield{author}{\bibinfo{person}{Chi Su}, \bibinfo{person}{Shiliang Zhang},
  \bibinfo{person}{Junliang Xing}, \bibinfo{person}{Wen Gao}, {and}
  \bibinfo{person}{Qi Tian}.} \bibinfo{year}{2016}\natexlab{}.
\newblock \showarticletitle{Deep attributes driven multi-camera person
  re-identification}. In \bibinfo{booktitle}{\emph{ECCV}}.
\newblock


\bibitem[\protect\citeauthoryear{Sun, Shrivastava, Singh, and Gupta}{Sun
  et~al\mbox{.}}{2017}]%
        {SSSG17}
\bibfield{author}{\bibinfo{person}{Chen Sun}, \bibinfo{person}{Abhinav
  Shrivastava}, \bibinfo{person}{Saurabh Singh}, {and} \bibinfo{person}{Abhinav
  Gupta}.} \bibinfo{year}{2017}\natexlab{}.
\newblock \showarticletitle{Revisiting unreasonable effectiveness of data in
  deep learning era}. In \bibinfo{booktitle}{\emph{ICCV}}.
\newblock


\bibitem[\protect\citeauthoryear{Thomee, Shamma, Friedland, Elizalde, Ni,
  Poland, Borth, and Li}{Thomee et~al\mbox{.}}{2015}]%
        {TSF+15}
\bibfield{author}{\bibinfo{person}{Bart Thomee}, \bibinfo{person}{David~A
  Shamma}, \bibinfo{person}{Gerald Friedland}, \bibinfo{person}{Benjamin
  Elizalde}, \bibinfo{person}{Karl Ni}, \bibinfo{person}{Douglas Poland},
  \bibinfo{person}{Damian Borth}, {and} \bibinfo{person}{Li-Jia Li}.}
  \bibinfo{year}{2015}\natexlab{}.
\newblock \showarticletitle{The new data and new challenges in multimedia
  research}.
\newblock \bibinfo{journal}{\emph{arXiv:1503.01817}} (\bibinfo{year}{2015}).
\newblock


\bibitem[\protect\citeauthoryear{Veit, Nickel, Belongie, and van~der
  Maaten}{Veit et~al\mbox{.}}{2018}]%
        {VNBV18}
\bibfield{author}{\bibinfo{person}{Andreas Veit}, \bibinfo{person}{Maximilian
  Nickel}, \bibinfo{person}{Serge Belongie}, {and} \bibinfo{person}{Laurens
  van~der Maaten}.} \bibinfo{year}{2018}\natexlab{}.
\newblock \showarticletitle{Separating self-expression and visual content in
  hashtag supervision}. In \bibinfo{booktitle}{\emph{CVPR}}.
\newblock


\bibitem[\protect\citeauthoryear{Wang, Cheng, and Feris}{Wang
  et~al\mbox{.}}{2016}]%
        {WCF16}
\bibfield{author}{\bibinfo{person}{Jing Wang}, \bibinfo{person}{Yu Cheng},
  {and} \bibinfo{person}{Rogerio~Schmidt Feris}.}
  \bibinfo{year}{2016}\natexlab{}.
\newblock \showarticletitle{Walk and learn: Facial attribute representation
  learning from egocentric video and contextual data}. In
  \bibinfo{booktitle}{\emph{CVPR}}.
\newblock


\bibitem[\protect\citeauthoryear{Wang, Zheng, Yang, Luo, and Tang}{Wang
  et~al\mbox{.}}{2019}]%
        {WZY+19}
\bibfield{author}{\bibinfo{person}{Xiao Wang}, \bibinfo{person}{Shaofei Zheng},
  \bibinfo{person}{Rui Yang}, \bibinfo{person}{Bin Luo}, {and}
  \bibinfo{person}{Jin Tang}.} \bibinfo{year}{2019}\natexlab{}.
\newblock \showarticletitle{Pedestrian attribute recognition: A survey}.
\newblock \bibinfo{journal}{\emph{arXiv:1901.07474}} (\bibinfo{year}{2019}).
\newblock


\bibitem[\protect\citeauthoryear{Xie, Girshick, Doll{\'a}r, Tu, and He}{Xie
  et~al\mbox{.}}{2017}]%
        {XGD+17}
\bibfield{author}{\bibinfo{person}{Saining Xie}, \bibinfo{person}{Ross
  Girshick}, \bibinfo{person}{Piotr Doll{\'a}r}, \bibinfo{person}{Zhuowen Tu},
  {and} \bibinfo{person}{Kaiming He}.} \bibinfo{year}{2017}\natexlab{}.
\newblock \showarticletitle{Aggregated residual transformations for deep neural
  networks}. In \bibinfo{booktitle}{\emph{CVPR}}.
\newblock


\bibitem[\protect\citeauthoryear{Yan, Yang, Sohn, and Lee}{Yan
  et~al\mbox{.}}{2016}]%
        {YYSL16}
\bibfield{author}{\bibinfo{person}{Xinchen Yan}, \bibinfo{person}{Jimei Yang},
  \bibinfo{person}{Kihyuk Sohn}, {and} \bibinfo{person}{Honglak Lee}.}
  \bibinfo{year}{2016}\natexlab{}.
\newblock \showarticletitle{Attribute2image: Conditional image generation from
  visual attributes}. In \bibinfo{booktitle}{\emph{ECCV}}.
\newblock


\bibitem[\protect\citeauthoryear{Yu and Grauman}{Yu and Grauman}{2014}]%
        {YG14}
\bibfield{author}{\bibinfo{person}{Aron Yu} {and} \bibinfo{person}{Kristen
  Grauman}.} \bibinfo{year}{2014}\natexlab{}.
\newblock \showarticletitle{Fine-grained visual comparisons with local
  learning}. In \bibinfo{booktitle}{\emph{CVPR}}.
\newblock


\bibitem[\protect\citeauthoryear{Zhang, Kyaw, Chang, and Chua}{Zhang
  et~al\mbox{.}}{2017}]%
        {ZKCC17}
\bibfield{author}{\bibinfo{person}{Hanwang Zhang}, \bibinfo{person}{Zawlin
  Kyaw}, \bibinfo{person}{Shih-Fu Chang}, {and} \bibinfo{person}{Tat-Seng
  Chua}.} \bibinfo{year}{2017}\natexlab{}.
\newblock \showarticletitle{Visual translation embedding network for visual
  relation detection}. In \bibinfo{booktitle}{\emph{CVPR}}.
\newblock


\bibitem[\protect\citeauthoryear{Zhao, Fu, Liang, Wu, Wang, and Wang}{Zhao
  et~al\mbox{.}}{2019}]%
        {ZFL+19}
\bibfield{author}{\bibinfo{person}{Bo Zhao}, \bibinfo{person}{Yanwei Fu},
  \bibinfo{person}{Rui Liang}, \bibinfo{person}{Jiahong Wu},
  \bibinfo{person}{Yonggang Wang}, {and} \bibinfo{person}{Yizhou Wang}.}
  \bibinfo{year}{2019}\natexlab{}.
\newblock \showarticletitle{A large-scale attribute dataset for zero-shot
  learning}. In \bibinfo{booktitle}{\emph{CVPRW}}.
\newblock


\end{thebibliography}

\end{document}